\documentclass[10pt,twocolumn,letterpaper]{article}

\usepackage{iccv}              % To produce the CAMERA-READY version
\usepackage{multicol,multirow}
\usepackage[utf8]{inputenc}
\usepackage{amsmath}
\usepackage{algorithm}
\usepackage{algorithmic}
\newcommand{\argmax}{\mathop{\mathrm{arg\,max}}}

\definecolor{iccvblue}{rgb}{0.21,0.49,0.74}
\usepackage[pagebackref,breaklinks,colorlinks,allcolors=iccvblue]{hyperref}

\def\paperID{11560} % *** Enter the Paper ID here
\def\confName{ICCV}
\def\confYear{2025}

\title{Enhancing Adversarial Transferability via Component-Wise Transformation}

\author{
Hangyu Liu\textsuperscript{1,2}\thanks{This work was done while Hangyu Liu was an intern at Westlake Milab.}, 
Bo Peng\textsuperscript{2}, 
Can Cui\textsuperscript{1},
Pengxiang Ding\textsuperscript{1,3}, 
Donglin Wang\textsuperscript{1}\thanks{Corresponding author.}
\\
\textsuperscript{1}Westlake University, \textsuperscript{2}Beijing University of Posts and Telecommunications, \textsuperscript{3}Zhejiang University
}

\begin{document}
\maketitle
\begin{abstract}

Deep Neural Networks (DNNs) are highly vulnerable to adversarial examples, which pose significant challenges in security-sensitive applications. Among various adversarial attack strategies, input transformation-based attacks have demonstrated remarkable effectiveness in enhancing adversarial transferability. However, existing methods still perform poorly across different architectures, even though they have achieved promising results within the same architecture. This limitation arises because, while models of the same architecture may focus on different regions of the object, the variation is even more pronounced across different architectures. Unfortunately, current approaches fail to effectively guide models to attend to these diverse regions. To address this issue, this paper proposes a novel input transformation-based attack method, termed Component-Wise Transformation (CWT). CWT applies interpolation and selective rotation to individual image blocks, ensuring that each transformed image highlights different target regions, thereby improving the transferability of adversarial examples. Extensive experiments on the standard ImageNet dataset show that CWT consistently outperforms state-of-the-art methods in both attack success rates and stability across CNN- and Transformer-based models.
\end{abstract}    
\section{Introduction}
\label{sec:intro}

With the rapid development of deep learning, artificial intelligence has achieved significant progress in computer vision, finding widespread applications in tasks such as image classification~\cite{he2016deep,hu2018squeeze}, object detection~\cite{ren2016faster,redmon2018yolov3}, and semantic segmentation~\cite{ronneberger2015u,long2015fully}. However, the adversarial vulnerability of deep learning models has gradually surfaced as a critical issue, limiting their deployment in real-world scenarios. Research shows that carefully crafted adversarial examples—created by adding imperceptible perturbations to normal input data—can cause deep learning models to make erroneous predictions~\cite{szegedy2013intriguing,goodfellow2014explaining,gowal2019scalable}. This phenomenon not only affects model performance on standard test datasets but poses severe risks in high-stakes applications~\cite{eykholt2018robust,yuan2022natural,sharif2016accessorize} such as autonomous driving and medical diagnostics.

\begin{figure}[!t]
    \centering
    % Uncomment and update the path to include the correct image file
    % \includegraphics[width=7.5cm]{figures/pos_attention_head.png}
    \includegraphics[width=8cm]{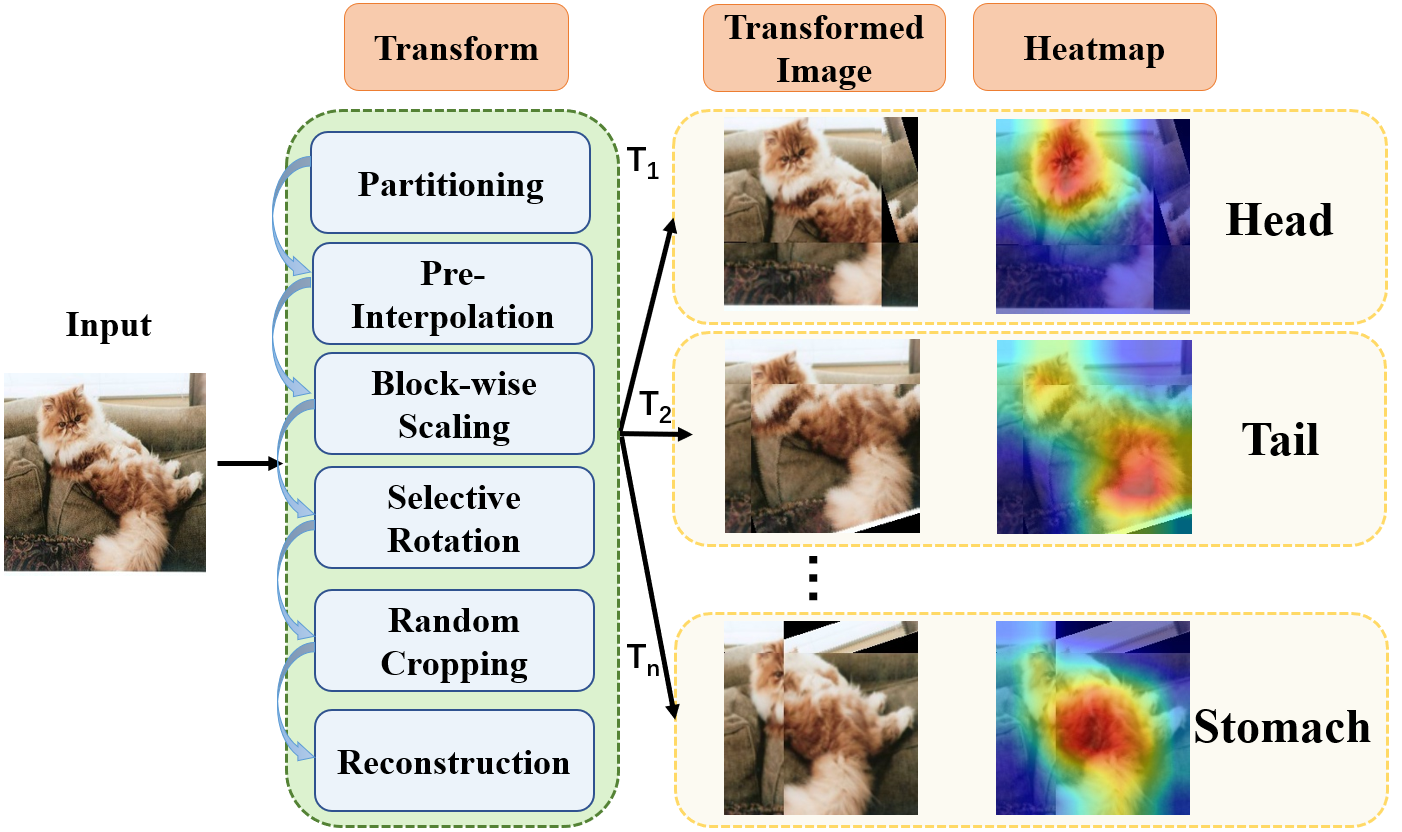}
    \caption{There shows our \textbf{CWT process} for image transformation. The heatmaps generated on ResNet-18. Our method successfully enables a single surrogate model to focus on different regions of an object.}
    % \vspace{-0.5em}
    \vspace{-1.5em}
    \label{fig:pipline}
\end{figure}

% \begin{figure}[!h]
%     \centering
%     % Uncomment and update the path to include the correct image file
%     % \includegraphics[width=7.5cm]{figures/pos_attention_head.png}
%     \includegraphics[width=8cm]{figtex/figures/resnet.png}
%     \caption{The heatmaps generated on ResNet-18 of raw image and its \textbf{transformed images} by DIM, SIM, Adimx, MaskBlock, US-MM, BSR, and our proposed CWT.}
%     \label{fig:transformed-images}
% \end{figure}

 % One critical property of adversarial attacks is transferability~\cite{xie2019improving,wang2021enhancing}, which refers to the ability of adversarial examples generated for one model (the surrogate model) to deceive other models (the target models), even when the target models have different architectures or training datasets. Transferability is essential in real-world black-box attack scenarios~\cite{liu2024differentially}, where attackers lack access to target model details and rely on surrogate models to craft adversarial examples. Enhancing transferability is a central challenge in adversarial attack research, as it determines the practical impact of adversarial examples.

 In black-box scenarios, attackers lack access to the details of the target models and can only rely on a limited number of surrogate models to craft adversarial examples. Therefore, a critical challenge is transferability, which refers to the ability of adversarial examples generated on surrogate models to deceive other target models, even when the target models have different architectures or are trained on different datasets~\cite{xie2019improving,wang2021enhancing}.

To improve adversarial transferability, researchers have proposed a variety of methods for generating adversarial examples. Gradient-based methods~\cite{goodfellow2014explaining,kurakin2018adversarial,dong2018boosting,lin2019nesterov,gao2020patch}  leverage the gradient information of models to craft adversarial examples with high computational efficiency and serve as the foundation for many other approaches. However, they generally suffer from poor transferability, particularly across models with different architectures. Model-related methods~\cite{wu2020skip,wei2022towards,ma2025improving} exploit internal model features, such as skip connections or backpropagation, to generate architecture-specific adversarial examples. While these methods perform well for specific architectures, they often overfit and lack generalizability. Ensemble-based~\cite{liu2016delving,xiong2022stochastic,li2023making,chen2023adaptive} and generation-based~\cite{salzmann2021learning} methods aggregate the outputs of multiple models or use generative models like GANs to directly generate adversarial examples. While these approaches can improve transferability, they are constrained by high computational costs. In contrast, input transformation-based methods~\cite{xie2019improving,xie2019improving,lin2019nesterov,zhang2023improving} enhance cross-model transferability through enriching adversarial example diversity with operations such as cropping, scaling, and rotation. These methods are computationally efficient and do not require access to model parameters, making them particularly suitable for black-box attack scenarios.

\begin{figure}[!t]
    \centering
    % Uncomment and update the path to include the correct image file
    % \includegraphics[width=7.5cm]{figures/pos_attention_head.png}
    \includegraphics[width=8cm]{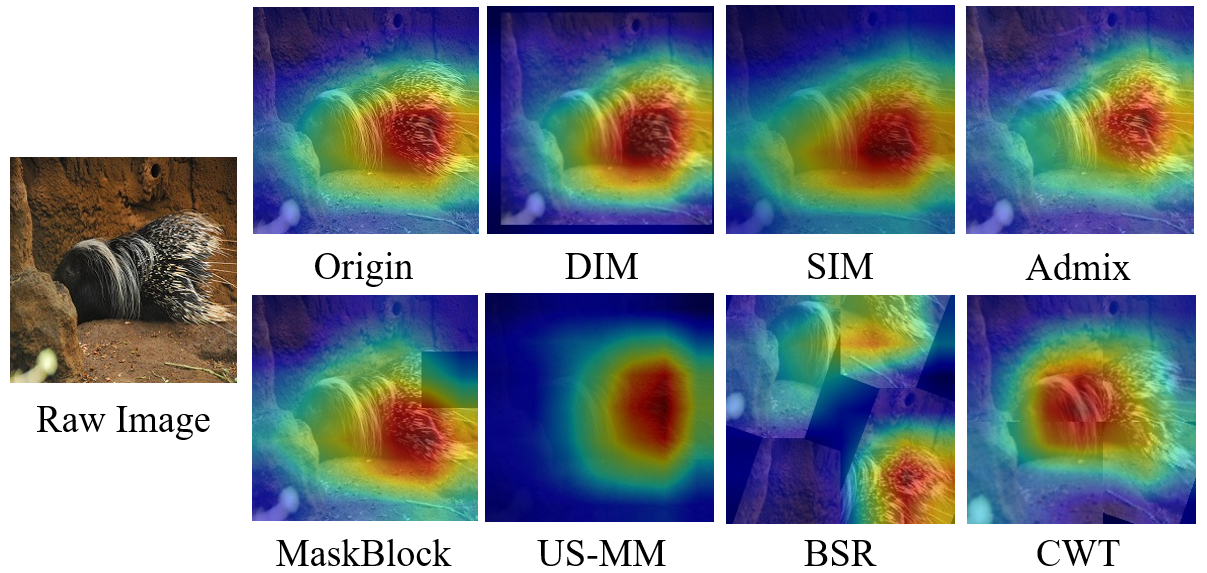}
    \caption{The heatmaps generated on ResNet-18 of raw image and its \textbf{transformed images} by DIM, SIM, Adimx, MaskBlock, US-MM, BSR, and our proposed CWT. Notably, CWT successfully shifts the surrogate model's attention toward the central region of the object.}
    \vspace{-1.0em}
    \label{fig:transformed-images}
\end{figure}

However, existing input transformation-based methods still face critical challenges. As illustrated in Figure \ref{fig:transformed-images}, some approaches fail to effectively alter the model's attention, resulting in an attention distribution that remains consistent with the original image. In contrast, others misdirect attention to irrelevant regions of the object, deviating from the areas of interest in the original image.

% However, existing input transformation-based methods still face challenges, 如图\ref{fig:transformed-images}所示，有的方法无法有效转移模型的注意力，有的方法把注意力转移到了对象无关的区域并引入了大量的信息损失

To address these issues, we propose a novel method called Component-Wise Transformation (CWT), which enhances the transferability of adversarial examples by encouraging models to focus on diverse regions of the object in the original image. Specifically, CWT applies block-wise transformations that interpolate and selectively rotate image patches while preserving essential semantic information. By introducing these localized transformations, our method generates adversarial examples with enriched attention distributions that generalize effectively across different model architectures.

Our contributions can be summarized as follows:
\begin{enumerate}
    \item We introduce a novel perspective on addressing the challenge of untargeted adversarial transferability. Instead of broadly enriching input images, our approach focuses on generating transformed images that enable a single surrogate model to attend to different regions of the object across various transformed images.
    
    \item We propose CWT, an innovative method combining interpolation and selective rotation to enhance adversarial transferability by diversifying attention distributions.
    
    \item  Our method achieves state-of-the-art results on the ImageNet dataset, demonstrating superior attack success rates and lower standard deviations compared to existing approaches.
\end{enumerate}

\section{Related Work}

 \begin{figure*}[!t]
\begin{center}
\vspace{-1.0em}
  \includegraphics[width=\textwidth]{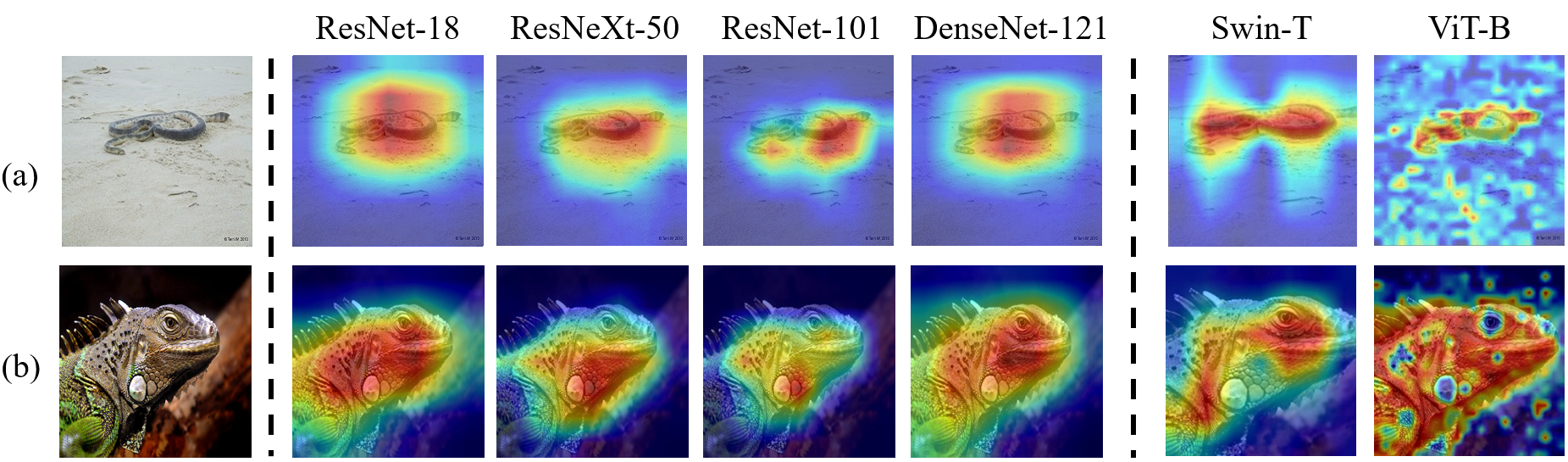} %插入图片，[]中设置图片大小，{}中是图片文件
\end{center}
\vspace{-1.0em}
\caption{\textbf{Demonstration of the different discriminative regions of the different models.} We adopt Grad-CAM~\cite{selvaraju2016grad} to visualize the attention maps of four CNN-based models—ResNet-18, ResNext-50, ResNet-101, DenseNet-121 and two Transformer-base models-Swin,Vit. }
\label{fig:attention-distribution}
% \vspace{-5mm}
\vspace{-1.0em}
\end{figure*}

\subsection{Input Transformation-based Attacks}

Input transformation-based methods have been widely studied to enhance adversarial transferability, particularly in black-box settings. These methods diversify the input samples used for gradient computation, enabling adversarial examples to generalize across different model architectures. We classify these methods into four categories: scale-based methods, mixed image-based methods, block-based methods, and adaptive-based methods. Scale-based methods~\cite{xie2019improving,lin2019nesterov} focus on resizing and transforming input images to introduce diversity in gradient computation. Mixed image-based methods~\cite{wang2021admix,wang2023boost,zhang2023improving} enrich input diversity by incorporating features from other images. However, whole-image transformations alone may not effectively alter the model's attention. To address this, Block-based methods~\cite{fan2022maskblock,wang2024boosting} focus on partitioning input images into blocks and applying transformations to each block independently. Furthermore, to better utilize existing transformation methods, adaptive-based methods~\cite{wei2023boosting,wang2023structure,zhu2024learning} introduce learnable transformation strategies to optimize adversarial transferability.

\subsection{Adversarial Defense}
Various defense approaches have been proposed to mitigate the threat of adversarial attacks, including adversarial training~\cite{madry2017towards,wang2021multi}, 
 feature denoising~\cite{xie2019feature}, certified defenses~\cite{raghunathan2018certified,gowal2019scalable} and  diffusion purification~\cite{wang2023better}. Among these, AT~\cite{shafahi2019adversarial} is one of the most effective methods, where adversarial examples are incorporated during training to enhance model robustness.   HGD~\cite{liao2018defense} utilizes a denoising autoencoder guided by high-level representations to eliminate adversarial perturbations. Similarly, NRP~\cite{naseer2020self}, leverages a self-supervised adversarial training mechanism to purify input samples, demonstrating strong effectiveness against transfer-based attacks. Certified defense methods, such as RS~\cite{cohen2019certified}, train robust classifiers by adding noise to the input and providing provable robustness within a certain radius. Furthermore, diffusion purification such as Diffure~\cite{nie2022diffusion} employs diffusion models to process input samples and reduce adversarial perturbations, further enhancing model resilience against adversarial threats.

% \input{figtex/five_tasks}

% 在本节中，我们首先介绍前置知识和我们的动机。然后我们将提供对于我们的BAC方法的细致解释，接着，为了突出与其他方法的不同，我们将总结我们的方法与SIM 以及 BSR 的不同。

\section{Methodology}

In this section, we first introduce the prerequisite knowledge and our motivation. Then, we provide a detailed explanation of our CWT method. To highlight the differences between our approach and other methods, we also summarize the distinctions between our method and DIM~\cite{xie2019improving}, as well as BSR~\cite{wang2024boosting}.

\subsection{Preliminaries}

Given a target model \( f \) with parameters \( \theta \) and a clean input \( x \) with the ground-truth label \( y \), the objective of an untargeted attacker is to generate an adversarial example \( x^{adv} \) that is visually similar to \( x \), such that \( \|x^{adv} - x\|_p \leq \epsilon \) and causes the model to misclassify the input, meaning \( f(x^{adv}; \theta) \neq y \). Here, \( \epsilon \) denotes the perturbation budget, and \( \| \cdot \|_p \) is the \( \ell_p \)-norm that quantifies the allowed perturbation. In this paper, we focus on the \( \ell_\infty \)-norm (where \( p = \infty \)), which restricts the maximum change to each individual pixel.

To generate such an adversarial example, the attacker generally maximizes the loss function \( J(x^{adv}, y; \theta) \) (e.g., cross-entropy loss) with respect to \( x^{adv} \), subject to the constraint \( \|x^{adv} - x\|_p \leq \epsilon \). This can be formalized as follows:

\begin{equation}
x^{adv} = \argmax_{\|x^{adv} - x\|_p \leq \epsilon} J(x^{adv}, y; \theta)
\end{equation}

Many methods have been proposed to approximate this optimization. Among these, the Momentum Iterative Fast Gradient Sign Method (MI-FGSM) ~\cite{dong2018boosting} is particularly notable for enhancing the transferability of adversarial examples across diverse models. 
This method incorporates a momentum term into the gradient calculation to stabilize the perturbation direction across iterations, thereby increasing the probability of successful attacks on various models. The update rule for the accumulated gradient \( g_t \) is:
\begin{equation}
g_{t+1} = \mu \cdot g_t + \frac{\nabla_x J(x^{adv}_t, y; \theta)}{\|\nabla_x J(x^{adv}_t, y; \theta)\|_1}
\end{equation}
\begin{equation}
x^{adv}_{t+1} = x^{adv}_t + \alpha \cdot \text{sign}(g_{t+1})
\end{equation}
where \( \mu \) is the decay factor that controls the influence of previous gradients, and \( g_t \) is the accumulated gradient at iteration \( t \). The \( \ell_1 \)-normalization of the gradient helps maintain consistency in the perturbation direction across iterations, enhancing the transferability of the attack.

\subsection{Motivation}

Despite differences in architectures and parameters, models trained for the same task often exhibit similar decision boundaries due to a shared latent space~\cite{huh2024platonic}. Adversarial attacks leveraging input transformations aim to generate diverse perturbations, enabling gradient-based methods to effectively explore this shared latent space and identify vulnerabilities in model decision boundaries.

However, experiments reveal a critical limitation: adversarial examples that perform well on the surrogate model often fail to generalize effectively to target models. This observation raises an important question: 

\emph{Why does this discrepancy occur?}

% \begin{figure}[!h]
%     \centering
%     % Uncomment and update the path to include the correct image file
%     % \includegraphics[width=7.5cm]{figures/pos_attention_head.png}
%     \includegraphics[width=8cm]{figtex/figures/resnet.png}
%     \caption{The heatmaps generated on ResNet-18 of raw image and its transformed images by DIM, SIM, Adimx, MaskBlock, US-MM, BSR, and our proposed CWT.}
%     \label{fig:transformed-images}
% \end{figure}

% As illustrated in Figure~\ref{fig:attention-distribution}, we observe that even when classification results are identical, the attention distributions of different models vary significantly. For instance(a), ResNet18 关注于蛇的腹部当Swin关注蛇头与蛇尾 focuses on \textbf{Region A}, while model YY emphasizes \textbf{Region B}. Motivated by these findings, we hypothesize that the transferability of adversarial attacks improves when perturbations better cover the regions of interest across all models. This hypothesis aligns with the principle of leveraging gradients from multiple models to optimize interactive perturbations. 

As illustrated in Figure~\ref{fig:attention-distribution}, we observe that even when classification results are identical, the attention distributions of different models vary significantly. For instance, as shown in Figure~\ref{fig:attention-distribution} (a), ResNet-18 focuses on the belly of the snake, while Swin focuses on both the head and tail of the snake. Motivated by these findings, we hypothesize that the transferability of adversarial attacks improves when perturbations better cover the regions of interest across all models. This hypothesis aligns with the principle of leveraging gradients from multiple models to optimize interactive perturbations.

To investigate this hypothesis, we first evaluate the attention distributions of transformed images generated by various input-transformation-based attack methods, as shown in Figure~\ref{fig:transformed-images}. It reveals that methods such as DIM, SIM, Admix, Maskblock, and US-MM fail to significantly alter the attention distributions. While BSR enriches the attention regions, it often shifts the focus to irrelevant areas outside the objects of interest in the original image. These findings prompt a novel question: 

\emph{How can we generate transformed images that encourage models to focus on diverse regions of the objects within an image?}

\subsection{Component-Wise Transformation}

% \begin{figure}[!h]
%     \centering
%     % Uncomment and update the path to include the correct image file
%     % \includegraphics[width=7.5cm]{figures/pos_attention_head.png}
%     \includegraphics[width=8cm]{figtex/figures/pipline.png}
%     \caption{(a) Random Masking: Forces the model to focus on alternative regions, but leads to significant information loss.
% (b) Zoom In: Applies a global zoom-in operation to the entire image, which fails to change the relative size and cause displacement of the object from its original position.
% }
%     \label{fig:pipline}
% \end{figure}

Encouraging models to focus on diverse regions of the object within an image is a challenging task for several reasons: 1) First, models tend to concentrate their attention on the most salient features of an object. Deep learning models are optimized to identify the most discriminative parts that contribute significantly to classification, often neglecting other informative regions. 2) Second, the attention distribution is highly dependent on model architecture and training data. Different models may attend to different regions of the same object, making it difficult to design a unified approach that effectively diversifies attention across various models.

To address these challenges, a straightforward method is to apply random masking to the image, forcing the model to utilize other regions for prediction. However, this approach leads to significant information loss, as substantial parts of the image are obscured, potentially impairing the attack algorithm performance.

Therefore, we define our objectives as:

\begin{itemize}
    
    \item \textbf{Maximize the retention of original image information}: Ensure that essential features and semantics of the original image are preserved.
    \item \textbf{Maximize the diversity of model attention regions}: Generate multiple transformed images where the model's attention covers as much as possible the region where the object is located in the original image.
\end{itemize}

% \begin{itemize}
    
%     \item \textbf{Maximize the retention of original image information}: Ensure that essential features and semantics of the original image are preserved to maintain the model's performance.
%     \item \textbf{Maximize the diversity of model attention regions}: Generate multiple transformed images where the model's attention covers as much as possible the region \( S(i:j, i':j') \) where the object is located in the original image.
% \end{itemize}

% These objectives can be formalized as the following optimization problem:

% \begin{equation}
% \begin{aligned}
% & \underset{T}{\text{maximize}}
% & & \mathcal{C}\left( \sum_{k=1}^{N} A_k^{T(I)}, S \right) \\
% & \text{subject to}
% & & \text{Loss}_{\text{info}}(I, T(I)) \leq \epsilon,
% \end{aligned}
% \label{eq:optimization}
% \end{equation}

% where:
% \begin{itemize}
%     \item \( T \) represents a set of transformations applied to the image \( I \).
%     \item \( \mathcal{A}_k^{T(I)} \) denotes the attention map of the model on the \( k \)-th transformed image \( T_k(I) \).
%     \item \( S(i:j, i':j') \) is the region in the original image where the object is located.
%     \item \( \mathcal{C}\left( \mathcal{A}_k^{T(I)}, S \right) \) is a coverage function measuring how well the attention maps \( \mathcal{A}_k^{T(I)} \) covers the object region \( S \).
%     \item \( \text{Loss}_{\text{info}}(I, T(I)) \) quantifies the information loss due to transformation, with \( \epsilon \) being an acceptable threshold.
% \end{itemize}

Motivated by human perception~\cite{proulx2010size}, where people tend to focus on larger objects, previous work~\cite{zhang2025mllms} has also shown that models tend to pay attention to larger regions within an image. This insight suggests that we could apply a \textbf{zoom-in operation} to the image to enlarge the object, potentially shifting the model's attention. However, globally scaling up the image fails to effectively change the relative size differences between different parts of the object and can cause significant displacement of the object from its original position.

% globally scaling up the image does not effectively alter the relative differences between different parts of the object 以及会使得对象较大的偏离原来的位置.

% To overcome these limitations,  we propose CWT. This method applies a sequence of transformations—including  interpolation ,包括\textbf{pre-interpolation}, 和\textbf{block-wise scaling}, 以及\textbf{selective rotation}, to individual blocks of an image. These transformations enhance local feature diversity and improve model robustness against adversarial perturbations.

% To overcome these limitations, we propose CWT, which applies a sequence of transformations to individual blocks of an image. These transformations include interpolation（更多对于插值的解释参见附录\ref{appedix: More Analysis on Interpolation}）, comprising pre-interpolation (interpolation-based shrinking) and block-wise scaling (interpolation-based enlargement), as well as selective rotation. Together, these operations enhance local feature diversity and improve the model's robustness against adversarial perturbations.

To overcome these limitations, we propose \textbf{CWT}, which applies a sequence of transformations to individual blocks of an image. As shown in Figure~\ref {fig:pipline}, these transformations include \textbf{bilinear interpolation} (for more details on interpolation, refer to Appendix B), comprising \textbf{pre-interpolation} (interpolation-based shrinking) and \textbf{block-wise scaling} (interpolation-based enlargement), as well as \textbf{selective rotation}.

Formally, the CWT process involves the following steps:

\begin{enumerate}

    \item \textbf{Image Partitioning}: Divide the image \( x \) into a grid of \( n \times n \) non-overlapping blocks \( \{ B_{i,j} \} \), where \( i, j \in \{1, 2, \dots, n\} \).

    \item \textbf{Pre-Interpolation:}  
    For each block \( B_{i,j} \), apply \textbf{scaled down} to reduce redundancy and compress features. Specifically, \( B_{i,j} \) is scaled by a random factor \( s_{i,j} \), where:
    \begin{equation}
    s_{i,j} \sim \text{Uniform}[s_{\text{min}}, s_{\text{max}}],
    \end{equation}
    The scaled-down block \( B'_{i,j} \) is calculated as:
    \begin{equation}
    B'_{i,j} = \text{Interpolate}(B_{i,j}, \text{size}=(H'_{i,j}, W'_{i,j})),
    \end{equation}
    where:
    \begin{equation}
    H'_{i,j} = \lfloor H_{i,j} / s_{i,j} \rfloor, \quad W'_{i,j} = \lfloor W_{i,j} / s_{i,j} \rfloor.
    \end{equation}

    The effectiveness of the Pre-Interpolation step is demonstrated in Figure~\ref{fig:interpolation}, which shows how it contributes to reducing redundancy and improving adversarial robustness.

    \item \textbf{Block-wise Scaling:}  
    After pre-interpolation, each block is scaled up using the same scaling factor \( s_{i,j} \), focusing attention on specific features. The scaled-up block \( \hat{B}_{i,j} \) is computed as:
    \begin{equation}
    \hat{B}_{i,j} = \text{Interpolate}(B'_{i,j}, \text{size}=(H''_{i,j}, W''_{i,j})),
    \end{equation}
    where:
    \begin{equation}
    H''_{i,j} = \lfloor H_{i,j} \cdot s_{i,j} \rfloor, \quad W''_{i,j} = \lfloor W_{i,j} \cdot s_{i,j} \rfloor.
    \end{equation}

    \item \textbf{Selective Rotation:}  
    To further diversify the spatial representation of features and avoid the massive information loss caused by rotating all blocks, Selective Rotation is applied to a subset of blocks. Let \( N \) be the total number of blocks, and let \( \mathcal{R} \subseteq \{1, 2, \dots, N\} \) denote the indices of \( k \) randomly selected blocks for rotation (\( |\mathcal{R}| = k \)). For each selected block \( B_{i,j} \in \mathcal{R} \), a random rotation angle \( r_{i,j} \) is sampled from:
    \begin{equation}
    r_{i,j} \sim \text{Uniform}(-r_{\text{max}}, r_{\text{max}}),
    \end{equation}
    where \( r_{\text{max}} \) defines the allowable rotation range. The block is rotated around its center, and any padding is filled with zeros to maintain the original dimensions.

    \item \textbf{Random Cropping:}  
    After scaling and optional rotation, each block \( \hat{B}_{i,j} \) is cropped back to its original size \( H_{i,j} \times W_{i,j} \).

    \item \textbf{Image Reconstruction:}  
    After all transformations are applied to each block, the blocks are reassembled into a single transformed image:
    \begin{equation}
    R_k(I) = \bigcup_{i=1}^{n} \bigcup_{j=1}^{n} \tilde{B}_{i,j}.
    \end{equation}

\end{enumerate}

To stabilize gradient calculations during adversarial training with CWT, we compute the average gradient over \( N \) transformed images:
\begin{equation}
\bar{g} = \frac{1}{N} \sum_{i=1}^{N} \nabla_{x^{\text{adv}}} J(T(x^{\text{adv}}, n, N, s_{\text{max}}, k, r), y; \theta)
\label{eq:gradient_averaging}
\end{equation}

Where \( T(x^{\text{adv}}, n, N, s_{\text{max}}, k, r) \) represents the transformation applied to the augmented image, parameterized by the input image \( x^{\text{adv}} \), the number of blocks \( n \), the total number of transformed images \( N \), the maximum scaling factor \( s_{\text{max}} \) (with \( s_{\text{min}} = 1.0 \)), the maximum rotation angle \( r \), and the number of rotated blocks \( k \). \( J \) denotes the loss function (cross-entropy), and \( y \) is the ground truth label.

Here we integrate our CWT method into MI-FGSM, and summarize the algorithm in Algorithm~\ref{alg:CWT}.

\begin{algorithm}[!hbtp]
\caption{Component-Wise Transformation}
\label{alg:CWT}
\begin{algorithmic}[1]
\setlength{\itemsep}{0pt} % 减少步骤间的垂直间距
\REQUIRE A classifier \( f \) with parameters \( \theta \), loss function \( J \),
a raw example \( x \) with ground-truth label \( y \),
the magnitude of perturbation \( \epsilon \), learning rate \( \alpha \), decay factor \( \mu \), number of iterations \( T \), number of transformed images \( N \), number of blocks \( n \), the maximum scaling factor \( s_{\text{max}} \) , the maximum rotation angle \( r \), and the number of rotated blocks \( k \).

\ENSURE An adversarial example \( x^{\text{adv}} \).
\STATE Initialize \( \delta = 0 \), \( g_0 = 0 \), and \( \alpha = \epsilon / T \)
\FOR{\( t = 1 \to T \)}

    \STATE Calculate the gradient by Eq.~\ref{eq:gradient_averaging}.
    \STATE Update the momentum \( g_t \) by:
    \vspace{-2mm}
    \[
    g_t = \mu \cdot g_{t-1} + \frac{\bar{g}}{\|\bar{g}\|_1}
    \]
    \vspace{-5mm}
    \STATE Update the adversarial perturbation \( \delta \) by:
    \vspace{-2mm}
    \[
    \delta = \alpha \cdot \text{sign}(g_t)
    \]
    \vspace{-5mm}
    
    \STATE Update the adversarial example by: \\
    \vspace{-2mm}
    \[ x^{\text{adv}}_t = x^{\text{adv}}_{t-1} + \delta 
    \]
    \vspace{-5mm}
    \STATE Clip \( \delta \) to ensure \( \|\delta\|_\infty \leq \epsilon \):
    \vspace{-2mm}
    \[  x^{\text{adv}}_t = \text{Clip}_{x, \epsilon}(x^{\text{adv}}_t - x, -\epsilon, \epsilon) \]
\ENDFOR
\RETURN $x^{\text{adv}}_T$
\end{algorithmic}
\end{algorithm}

% \begin{algorithm}[H]
% \caption{Block Enlarging and Low-loss Rotation }
% \label{alg:CWT}
% \begin{algorithmic}[1]
% \setlength{\itemsep}{0pt} % 减少步骤间的垂直间距
% \REQUIRE A classifier \( f \) with parameters \( \theta \), loss function \( J \); \\
% a raw example \( x \) with ground-truth label \( y \); \\
% the magnitude of perturbation \( \epsilon \); \\
% number of iterations \( T \); decay factor \( \mu \); \\
% number of transformed images \( N \); number of blocks \( n \); learning rate \( \alpha \).
% \ENSURE An adversarial example \( x^{\text{adv}} \).
% \STATE Initialize \( \delta = 0 \), \( g_0 = 0 \), and \( \alpha = \epsilon / T \)
% \FOR{\( t = 1 \to T \)}
%     \STATE Apply transformations using CWT: \( T(x + \delta, n, N) \)
%     \STATE Compute the logits \( \text{logits} = f(T(x + \delta, n, N), \theta) \)
%     \STATE Compute the loss:
%     \[
%     \text{loss} = J(\text{logits}, y)
%     \]
%     \STATE Calculate the gradient:
%     \[
%     \text{grad} = \nabla_\delta J(\text{logits}, y)
%     \]
%     \STATE Update the momentum \( g_t \) by:
%     \[
%     g_t = \mu \cdot g_{t-1} + \frac{\text{grad}}{\|\text{grad}\|_1}
%     \]
%     \STATE Update the adversarial perturbation \( \delta \) by:
%     \[
%     \delta = \delta + \alpha \cdot \text{sign}(g_t)
%     \]
%     \STATE Clip \( \delta \) to ensure \( \|\delta\|_\infty \leq \epsilon \)
% \ENDFOR
% \RETURN \( x^{\text{adv}} = x + \delta \)
% \end{algorithmic}
% \end{algorithm}

\subsection{Comparison of CWT with DIM and BSR}

DIM enhances the transferability of adversarial attacks by applying resizing and padding to input samples, whereas BSR achieves this by randomly shuffling and rotating image blocks. Despite incorporating similar operations, our method, CWT, differs significantly from these approaches. Below, we highlight the distinctions between our method and the two methods.

\begin{itemize}
    \item \textbf{CWT Vs. DIM} In contrast to DIM, which performs scaling with fixed values on the entire image, our method emphasizes a random enlargement operation with pre-interpolation. Moreover, while DIM applies this operation to the entire image uniformly, our approach targets individual blocks within the image, introducing more localized transformations.

    \item \textbf{CWT Vs. BSR} As for BSR, although both methods involve block partitioning, our approach discards the shuffle operation and primarily utilizes block interpolation to modify the image structure. Regarding rotation, instead of applying rotation to all blocks, our method adopts a selective rotation mechanism, where the number of rotated blocks is limited.
\end{itemize}

\section{Experiments}

\begin{table*}[h!]
    \centering
    \scriptsize
    \setlength{\tabcolsep}{3pt}
    \renewcommand{\arraystretch}{0.7}
    \resizebox{\textwidth}{!}{
        \begin{tabular}{c c c c c c c c c c c c}
            \toprule
            \textbf{Model} & \textbf{Attack} & \textbf{RN-18 ($\uparrow$)} & \textbf{RN-101 ($\uparrow$)} & \textbf{RX-50 ($\uparrow$)} & \textbf{DN-121 ($\uparrow$)} & \textbf{ViT-B ($\uparrow$)} & \textbf{PiT-B ($\uparrow$)} & \textbf{ViF-S ($\uparrow$)} & \textbf{Swin-T ($\uparrow$)} & \textbf{Mean ($\uparrow$)} & \textbf{Std. Dev. ($\downarrow$)}\\
            \midrule
            \multirow{7}{*}{\textbf{RN-18}} 
              & DIM   & \textbf{100*} & 61.7          & 66.1           & 90.4          & 30.4           & 37.4           & 53.4          & 56.9           & 62.0          & 23.8          \\
 & SIM   & \textbf{100*} & 59.6          & 64.1           & 90.5          & 24.6           & 35.7           & 49.0            & 53.3           & 59.6          & 25.5          \\
 & Admix & \textbf{100*} & 69.9          & 74.6           & 95.4          & 31.2           & 42.6           & 59.8          & 63.0             & 67.1          & 23.6          \\
  & MaskBlock & \textbf{100*} & 48.8          & 50.4          & 79.6          & 18.6          & 25.3          & 38.4          & 43.7          & 50.6          & 27.1          \\
 & US-MM & \textbf{100.0*} & 66.6 & 71.7 & 94.3 & 29.1 & 41.0 & 56.8 & 60.3 & 65.0 & 24.2 \\

 & BSR   & \textbf{100.0*} & 89.1          & 90.2           & \textbf{99.4}          & 49.1           & 62.3           & 79.4          & 79.2           & 81.1          & 16.7          \\
 & CWT  & \textbf{100.0*} & \textbf{90.2 }         & \textbf{93.7 }          & \textbf{99.4}          & \textbf{55.9}           & \textbf{68.8}           & \textbf{84.1}          & \textbf{83.6}           & \textbf{84.5}        & \textbf{14.3}          \\
          \midrule
            \multirow{7}{*}{\textbf{RN-101}} 
             & DIM   & 61.7          & 84.7*         & 63.0             & 65.5          & 30.2           & 40.9           & 48.0            & 48.2           & 55.3          & 16.9          \\
 & SIM   & 62.5          & 91.7*         & 64.7           & 66.3          & 25.8           & 38.1           & 46.3          & 46.1           & 55.2          & 20.4          \\
 & Admix & 74.7          & 94.9*         & 77.5           & 77.4          & 36.0             & 49.7           & 60.9          & 58.6           & 66.2          & 18.6          \\
 & MaskBlock & 61.3          & 92.3*         & 56.6          & 61.9          & 19.5          & 29.8          & 36.5          & 38.5          & 49.6          & 23.2          \\
& US-MM & 81.8 & \textbf{95.1*}& 80.1 & 83.7 & 35.3 & 50.7 & 62.7 & 60.6 & 68.8 & 19.9 \\

 & BSR   & 86.7          & 94.6* & 89.6           & 90.2          & 58.5           & 72.9           & 80.8          & 78.1           & 81.4          & 10.9         \\
 & CWT & \textbf{87.7}          & \textbf{95.1*} & \textbf{91.0}           & \textbf{91.6}          & \textbf{67.3}           & \textbf{78.5}           & \textbf{85.5}          & \textbf{81.9}           & \textbf{84.8}          & \textbf{8.3}          \\
            \midrule
            \multirow{7}{*}{\textbf{RX-50}} 
          & DIM   & 61.3          & 55.8          & 86.7*          & 63.0            & 26.0             & 35.2           & 45.7          & 44.4           & 52.3          & 18.8          \\
 & SIM   & 59.5          & 57.8          & 94.0*            & 64.6          & 20.9           & 32.9           & 39.7          & 41.6           & 51.4          & 22.7          \\
 & Admix & 71.8          & 72.9          & 95.8*          & 75.6          & 29.6           & 44.4           & 53.7          & 53.9           & 62.2          & 20.8          \\
& MaskBlock & 55.3          & 46.2          & 94.1*         & 54.5          & 16.4          & 24.2          & 32.5          & 33.1          & 44.5          & 24.4          \\
 & US-MM & 78.6 & 73.2 & \textbf{96.9*} & 82.1 & 31.5 & 46.8 & 57.6 & 55.6 & 65.3 & 21.2 \\

 & BSR   & 85.7          & 86.0          & 96.5* & 88.7          & 48.2           & 68.3           & 76.0          & 73.9           & 77.9          & 14.1          \\
 & CWT & \textbf{87.3}          & \textbf{86.4}          & 95.9* & \textbf{90.2}          & \textbf{57.7}           & \textbf{72.3}           & \textbf{80.9}          & \textbf{78.4}           & \textbf{81.1}          & \textbf{11.2}         \\
            \midrule
            \multirow{7}{*}{\textbf{DN-121}} 
               & DIM   & 86.7          & 70.7          & 72.0             & 99.9*         & 34.0             & 43.6           & 60.6          & 57.2           & 65.6          & 21.6          \\
 & SIM   & 90.0            & 70.3          & 72.8           & \textbf{100*} & 31.5           & 41.0             & 57.0            & 59.6           & 65.3          & 23.1          \\
 & Admix & 95.3          & 80.5          & 83.2           & \textbf{100*} & 39.0             & 51.6           & 68.7          & 68.7           & 73.4          & 20.8          \\
 & MaskBlock & 82.8          & 59.3          & 61.9          & \textbf{100*} & 23.8          & 34.4          & 47.9          & 49.3          & 57.4          & 24.8          \\
 & US-MM & 95.6 & 77.9 & 79.9 & 99.9* & 38.1 & 48.8 & 67.6 & 65.3 & 71.6 & 21.3 \\

 & BSR   & 98.0          & 89.7          & 92.2           & \textbf{100.0*} & 52.1           & 66.9           & 81.6          & 79.4           & 82.5          & 15.3        \\
 & CWT  & \textbf{99.0}          & \textbf{90.9}          & \textbf{93.8 }          & \textbf{100.0*} & \textbf{56.4}           & \textbf{70.8}           & \textbf{85.3}          & \textbf{82.2 }          & \textbf{84.8}         & \textbf{14.0}         \\
            \bottomrule
        \end{tabular}
    }
     \caption{Attack success rates (\%) on eight models under single model setting with various single input transformations. The surrogate models are \textbf{cnn-based}. * indicates the surrogate model.}
\label{tab:cnn-based}
\end{table*}

\subsection{Experimental Setup}

\textbf{Models.} We evaluate our proposed CWT method across three categories of target models: 1) \textbf{CNN-based models}, including four widely recognized architectures: ResNet-18 (RN-18)~\cite{he2016deep}, ResNet-101 (RN-101)~\cite{he2016deep}, ResNeXt-50 (RX-50)~\cite{xie2017aggregated}, and DenseNet-121 (DN-121)~\cite{huang2017densely}; 2) \textbf{Transformer-based models}, comprising ViT-B~\cite{dosovitskiy2020image}, PiT-B~\cite{heo2021rethinking}, Visformer (ViF-S)~\cite{chen2021visformer}, and Swin-Tiny (Swin-T)~\cite{liu2021swin}; 3) \textbf{Defense models}, which include four defense methods: AT~\cite{shafahi2019adversarial}, HGD~\cite{liao2018defense}, RS~\cite{cohen2019certified}, and  DiffPure~\cite{nie2022diffusion}; All models are pre-trained on the ImageNet dataset and evaluated on single model.

\textbf{Dataset.} Following previous works\footnote{\url{https://github.com/Trustworthy-AI-Group/TransferAttack}}, We evaluate our proposed CWT on 1000 images belonging to 1000 categories from the validation set of ImageNet dataset. All images are classified correctly by the models.

% \textbf{Baselines.}我们对比CWT与其他Input transformation-based方法，具体来说，基于图像尺度的方法DIM、SIM，基于混合图像的方法Admix、US-MM，基于分块的方法MaskBlock、BSR。For fairness, all the input transformations are integrated into MI-FGSM.

\textbf{Baselines.} We compare CWT with other input transformation-based methods. Specifically, the image scale-based methods (DIM~\cite{xie2019improving}, 
 SIM~\cite{lin2019nesterov}), the mixed image-based methods (Admix~\cite{wang2021admix}, US-MM~\cite{wang2023boost}), and the block-based methods (MaskBlock~\cite{fan2022maskblock}, BSR~\cite{wang2024boosting}). For fairness, all the input transformations are integrated into MI-FGSM~\cite{dong2018boosting}.

\textbf{Evaluation Settings.} We set the maximum perturbation $\epsilon = 16/255$, the number of iterations $epoch = 10$, the step size $\alpha = \epsilon / epoch$, the batch size $batchsize = 8$, and the decay factor $\mu = 1$ for MI-FGSM. For our method, CWT generates 20 scaled copies per iteration, divides the image into 2x2 blocks, applies a scaling factor ranging from 1.0 to 1.3, and applies a maximum rotation angle of $26^\circ$, selectively rotating $k = 2$ blocks. For other methods, we follow the parameters reported in the original papers.

\subsection{Evaluations on CNN-based Models}

We first evaluate the transferability of adversarial examples generated by various attacks with input transformations when the surrogate model is based on a CNN architecture. Specifically, we generate adversarial examples using RN-18, RN-101, RX-50, and DN-121 as surrogate models and evaluate their attack success rates (ASR) across eight target models. The results are summarized in Table~\ref{tab:cnn-based}, where the first column indicates the surrogate model, the second column lists the attack methods, and the remaining columns present the ASR of different classification models under attack and the last two columns are the mean ASR and standard deviation, respectively.

Our proposed method consistently achieves state-of-the-art performance, with the highest mean ASR and lowest standard deviation across almost all experimental settings.  Notably, when the surrogate model is RN-18, our method achieves the highest ASR across all target models. Furthermore, when the target models are ViT-B and PiT-B, our method surpasses the previous state-of-the-art BSR more than 5\% and outperforms other baseline methods by at least 17.4\%. These results confirm the effectiveness of our approach in generating transferable adversarial examples when the surrogate model is CNN-based.

\subsection{Evaluations on Transformer-based models}

\begin{table*}[h!]
    \centering
    \scriptsize
    \setlength{\tabcolsep}{3pt}
    \renewcommand{\arraystretch}{0.7}
    \resizebox{\textwidth}{!}{
        \begin{tabular}{c c c c c c c c c c c c}
            \toprule
            \textbf{Model} & \textbf{Attack} & \textbf{RN-18 ($\uparrow$)} & \textbf{RN-101 ($\uparrow$)} & \textbf{RX-50 ($\uparrow$)} & \textbf{DN-121 ($\uparrow$)} & \textbf{ViT-B ($\uparrow$)} & \textbf{PiT-B ($\uparrow$)} & \textbf{ViF-S ($\uparrow$)} & \textbf{Swin-T ($\uparrow$)} & \textbf{Mean ($\uparrow$)} & \textbf{Std. Dev. ($\downarrow$)}\\
            \midrule
            \multirow{7}{*}{\textbf{ViT-B}}
               & DIM   & 56.9          & 47.8          & 49.6           & 59.7          & 89.5*          & 54.0             & 55.0            & 61.7           & 59.3          & 13.1          \\
 & SIM   & 62.4          & 48.6          & 52.4           & 64.3          & \textbf{99.1*} & 57.8           & 58.7          & 71.9           & 64.4          & 15.7          \\
 & Admix & 65.6          & 52.4          & 55.3           & 65.9          & 98.9*          & 61.3           & 63.5          & 73.1           & 67.0          & 14.4          \\
 & MaskBlock & 59.3          & 41.4          & 44.1          & 59.1          & 98.9*         & 46.9          & 51.3          & 62.6          & 58.0          & 18.3          \\
 & US-MM & 69.2 & 54.7 & 57.6 & 68.7 & 97.4* & 61.8 & 65.0 & 75.9 & 68.8 & 13.4 \\

  &BSR   & 78.2 & 74.7 & 75.9 & 81.0 & 90.2* & 81.5 & 79.2 & 81.6 & 80.3 & \textbf{4.4} \\
              & CWT   & \textbf{78.3} & \textbf{75.3} & \textbf{77.2} & \textbf{81.6} & 93.1* & \textbf{83.5} & \textbf{82.4} & \textbf{83.9} & \textbf{81.9} & 5.1 \\
            \midrule
            \multirow{7}{*}{\textbf{PiT-B}} 
              & DIM   & 59.5          & 50.5          & 54.5           & 62.2          & 47.9           & 91.6*          & 63.5          & 65.3           & 61.9          & 13.5          \\
 & SIM   & 58.6          & 45.4          & 48.7           & 60.1          & 38.3           & 97.7*          & 56.8          & 60.9           & 58.3          & 17.8          \\
 & Admix & 60.4          & 47.7          & 51.8           & 60.4          & 42.9           & 94.6*          & 61.1          & 63.5           & 60.3          & 15.7          \\
 & MaskBlock & 59.2          & 40.4          & 45.2          & 57.3          & 35.4          & \textbf{99.2*}         & 55.9          & 57.2          & 56.2          & 19.5          \\
 & US-MM & 66.1 & 54.1 & 56.4 & 64.8 & 45.0 & 93.9* & 64.0 & 68.0 & 64.0 & 14.3 \\

 & BSR   & 82.9 & 80.8 & 84.0 & 86.5 & 75.0 & 97.8* & 90.2 & 90.4 & 86.0 & 6.5 \\
              & CWT   & \textbf{85.8} & \textbf{85.1} & \textbf{87.3} & \textbf{90.7} & \textbf{85.3} & 97.8* & \textbf{92.6} & \textbf{92.6} & \textbf{90.0} & \textbf{4.3} \\
            \midrule
            \multirow{7}{*}{\textbf{ViF-S}} 
              & DIM   & 71.2          & 63.7          & 67.1           & 75.2          & 53.4           & 71.0             & 95.1*         & 76.9           & 71.7          & 12.0          \\
 & SIM   & 68.1          & 60.5          & 62.4           & 72.0            & 49.7           & 65.9           & 96.7*         & 74.8           & 68.8          & 13.7          \\
 & Admix & 75.1          & 67.0            & 70.6           & 78.3          & 56.5           & 72.8           & 97.0*           & 81.6           & 74.9          & 11.8          \\
  & MaskBlock & 64.2          & 47.3          & 51.3          & 64.4          & 35.7          & 54.9          & 99.2*         & 65.2          & 60.3          & 18.7          \\
 & US-MM & 82.1 & 70.5 & 73.6 & 83.3 & 56.2 & 73.7 & 97.6* & 82.9 & 77.5 & 12.1 \\

 & BSR   & \textbf{90.7} & 86.9 & \textbf{90.7} & 93.5 & 73.3  & 88.9 & \textbf{99.3*} & 92.7 & 89.5 & 7.0 \\
              & CWT   & 89.1 & \textbf{87.6} & 89.9 & \textbf{94.0} & \textbf{80.4} & \textbf{91.9} & 99.1* & \textbf{93.7} & \textbf{90.7} & \textbf{5.1} \\
            \midrule
            \multirow{7}{*}{\textbf{Swin-T}} 
              & DIM   & 67.2          & 53.4          & 56.8           & 68.8          & 48.7           & 65.3           & 69.3          & 96.2*          & 65.7          & 14.5          \\
 & SIM   & 48.9          & 29.9          & 34.1           & 45.4          & 27.7           & 35.5           & 43.1          & 98.1*          & 45.3          & 22.6          \\
 & Admix & 55.1          & 33.9          & 37.4           & 50.0            & 28.3           & 37.8           & 47.2          & 98.2*          & 48.5          & 22.0          \\
 & MaskBlock & 47.8          & 26.5          & 30.9          & 44.5          & 24.4          & 31.6          & 39.1          & 98.3*         & 42.9          & 23.9         \\
 & US-MM & 57.5 & 32.2 & 37.8 & 52.2 & 26.8 & 37.8 & 47.4 & 96.8* & 48.6 & 22.0 \\

 & BSR   & 88.7 & 82.5 & 86.0 & 91.2 & 71.6 & 89.7 & 91.0 & 98.3* & 87.4 & 7.3 \\
              & CWT   & \textbf{90.7} & \textbf{85.9} & \textbf{88.1} & \textbf{93.6} & \textbf{80.8} & \textbf{92.9} & \textbf{94.1} & \textbf{98.5*} & \textbf{90.6} & \textbf{5.2} \\
            \bottomrule
        \end{tabular}
    }
        \caption{Attack success rates (\%) on eight models under single model setting with various single input transformations. The surrogate models are \textbf{Transformer-based}. * indicates the surrogate model.}
\label{tab:transformer-based}
\end{table*}

\begin{table}[!h]
    \centering
    \normalsize % 调整字体大小
    \renewcommand{\arraystretch}{0.9} % 调整行间距
    \resizebox{0.48\textwidth}{!}{ % 将表格缩放到半栏宽度
        \begin{tabular}{c c c c c c c c c}
            \toprule
            \textbf{Attack} & \textbf{RN-18 ($\uparrow$)} & \textbf{RN-101 ($\uparrow$)} & \textbf{RX-50 ($\uparrow$)} & \textbf{DN-121 ($\uparrow$)} & \textbf{ViT-B ($\uparrow$)} & \textbf{PiT-B ($\uparrow$)} & \textbf{ViF-S ($\uparrow$)} & \textbf{Swin-T ($\uparrow$)} \\
            \midrule
            DIM & 36.4 & 32.9 & 32.0 & 34.6 & 33.4 & 33.2 & 33.5 & 33.7 \\
            SIM & 36.5 & 31.6 & 32.1 & 35.6 & 35.4 & 33.4 & 34.3 & 31.0 \\
            ADMIX & 38.4 & 32.6 & 32.2 & 38.0 & 35.9 & 33.5 & 34.3 & 31.8 \\
            MaskBlock & 34.0 & 31.7 & 31.3 & 32.6 & 33.7 & 32.6 & 32.2 & 31.0 \\
            US-MM & 38.3 & 34.9 & 33.7 & 37.9 & 36.4 & 34.8 & 35.4 & 32.0 \\
            BSR & 40.3 & 36.9  & 35.2  & 38.9 & \textbf{38.0}  &  35.1 & 36.7 &  37.5 \\
            CWT & \textbf{41.1} & \textbf{38.0} & \textbf{36.9 } & \textbf{39.9} &  37.7 & \textbf{36.9} & \textbf{37.7} & \textbf{39.4} \\
            \bottomrule
        \end{tabular}
    }
    \caption{Attack success rates (\%) of adversarial examples generated using various attack methods across eight classification models under \textbf{AT}.}
    \label{tab:AT}
\end{table}

\begin{table}[!h]
    \centering
    \normalsize % 调整字体大小
    \renewcommand{\arraystretch}{0.9} % 调整行间距
    \resizebox{0.48\textwidth}{!}{ % 将表格缩放到半栏宽度
        \begin{tabular}{c c c c c c c c c}
            \toprule
            \textbf{Attack} & \textbf{RN-18 ($\uparrow$)} & \textbf{RN-101 ($\uparrow$)} & \textbf{RX-50 ($\uparrow$)} & \textbf{DN-121 ($\uparrow$)} & \textbf{ViT-B ($\uparrow$)} & \textbf{PiT-B ($\uparrow$)} & \textbf{ViF-S ($\uparrow$)} & \textbf{Swin-T ($\uparrow$)} \\
            \midrule
            DIM & 24.4 & 29.8 & 20.7 & 27.2 & 25.7 & 23.4 & 26.8 & 21.1 \\
            SIM & 23.3 & 28.9 & 18.0 & 24.7 & 27.8 & 20.9 & 25.9 & 14.2 \\
            ADMIX & 24.2 & 35.2 & 23.3 & 30.5 & 28.9 & 23.1 & 29.3 & 14.9 \\
            MaskBlock & 18.8 & 22.7 & 15.5 & 20.2 & 23.5 & 21.8 & 21.1 & 12.9 \\
            US-MM & 24.5 & 33.8 & 23.2 & 27.0 & 30.1 & 23.3 & 29.4 & 13.8 \\

            BSR & 33.6 & 45.5 & 28.8 & 33.8 & 39.5 & 35.0 & 37.9 & 30.5 \\
            CWT & \textbf{40.5} & \textbf{56.7} & \textbf{38.4} & \textbf{41.3} & \textbf{43.2} & \textbf{43.2} & \textbf{48.7} & \textbf{37.8} \\
            \bottomrule
        \end{tabular}
    }
    \caption{Attack success rates (\%) of adversarial examples generated using various attack methods across eight classification models under \textbf{Diffure}. The classifier is RN-101.}
    \label{tab:diffure}
\end{table}

Building on the results observed with CNN-based models, we next evaluate our method's performance when the surrogate model is based on a transformer architecture. Specifically, we use Transformer-based models, including ViT-B, PiT-B, ViF-S, and Swin-T, as surrogate models to generate adversarial examples and evaluate their transferability. The results are summarized in Table~\ref{tab:transformer-based}.

From the results, we observe that our method also achieves SOTA performance in terms of both mean ASR and standard deviation across most of experimental settings. Notably, when the surrogate model is ViT-B, our method does not achieve the highest ASR on the original classification model (ViT-B itself) but consistently achieves the best performance on all other classification models. This demonstrates the robustness and generalizability of our approach, as it effectively transfers adversarial examples to different target architectures, even when the surrogate model has inherent challenges in transferability.

\subsection{Evaluations on Defense Method}

To comprehensively assess the robustness of our proposed method against diverse defense mechanisms, we conducted experiments under four defense settings: Adversarial Training (AT), High-level representation Guided Denoiser (HGD), Randomized Smoothing (RS), and Diffusion Purification 
 (Diffure). More extensive evaluations can be found in Appendix A. As summarized in Tables~\ref{tab:AT} through~\ref{tab:RS}, our method, particularly CWT, consistently achieves the highest attack success rates across most of defense settings, outperforming BSR and other baseline methods. For instance, under Diffure, it outperforms BSR by 8.1\% in mean ASR and exceeding weaker baselines by more than 17\%. Even under RS, a certified defense known for its robustness, our method maintains competitive ASRs (e.g., 28.7\% for RN-101 and 29.5\% for ViF-S), showcasing its resilience. These results validate the robustness and effectiveness of our approach across diverse defenses.

\subsection{Ablation Study}

To further evaluate CWT, we conduct ablation studies on five key hyperparameters and one critical step: \textit{number of blocks} $n$, \textit{maximum scale factor} $s_{\text{max}}$, \textit{maximum rotation angle} \( r \), \textit{number of rotated blocks} \( k \), \textit{number of transformed copies} $N$, and the key step \textit{pre-interpolation}. All experiments are conducted using RN-101 as the surrogate model.

\begin{table}[!h]
    \centering
    \normalsize % 调整字体大小
    \renewcommand{\arraystretch}{0.9} % 调整行间距
    \resizebox{0.48\textwidth}{!}{ % 将表格缩放到半栏宽度
        \begin{tabular}{c c c c c c c c c}
            \toprule
            \textbf{Attack} & \textbf{RN-18 ($\uparrow$)} & \textbf{RN-101 ($\uparrow$)} & \textbf{RX-50 ($\uparrow$)} & \textbf{DN-121 ($\uparrow$)} & \textbf{ViT-B ($\uparrow$)} & \textbf{PiT-B ($\uparrow$)} & \textbf{ViF-S ($\uparrow$)} & \textbf{Swin-T ($\uparrow$)} \\
            \midrule
            DIM & 56.9 & 49.7 & 42.1 & 66.7 & 43.0 & 45.7 & 57.2 & 46.2 \\
            SIM & 52.7 & 43.5 & 36.4 & 64.3 & 42.4 & 37.2 & 49.1 & 22.5 \\
            ADMIX & 63.1 & 58.2 & 50.6 & 76.5 & 46.8 & 39.9 & 57.2 & 25.9 \\
            MaskBlock & 38.2 & 34.0 & 26.8 & 48.9 & 35.9 & 32.3 & 36.9 & 20.5 \\
            US-MM & 59.4 & 60.1 & 52.2 & 72.5 & 48.2 & 44.8 & 61.2 & 24.5 \\

            BSR & 86.9 & 80.9 & 74.9 & 90.1 & 71.9 & 74.2 & 81.0 & 75.0 \\
            CWT & \textbf{91.2} & \textbf{85.4} & \textbf{81.9} & \textbf{93.1} & \textbf{73.9} & \textbf{81.8} & \textbf{85.0} & \textbf{83.5} \\
            \bottomrule
        \end{tabular}
    }
    \caption{Attack success rates (\%) of adversarial examples generated using various attack methods across eight classification models under \textbf{HGD}.}
    \label{tab:HGD}
\end{table}
\begin{table}[!h]
    \centering
    \normalsize % 调整字体大小
    \renewcommand{\arraystretch}{0.9} % 调整行间距
    \resizebox{0.48\textwidth}{!}{ % 将表格缩放到半栏宽度
        \begin{tabular}{c c c c c c c c c}
            \toprule
            \textbf{Attack} & \textbf{RN-18 ($\uparrow$)} & \textbf{RN-101 ($\uparrow$)} & \textbf{RX-50 ($\uparrow$)} & \textbf{DN-121 ($\uparrow$)} & \textbf{ViT-B ($\uparrow$)} & \textbf{PiT-B ($\uparrow$)} & \textbf{ViF-S ($\uparrow$)} & \textbf{Swin-T ($\uparrow$)} \\
            \midrule
            DIM & 26.0 & 21.9 & 22.0 & 25.2 & 23.6 & 22.2 & 22.9 & 22.9 \\
            SIM & 26.0 & 21.3 & 21.3 & 26.0 & 25.2 & 22.7 & 24.0 & 21.5 \\
            ADMIX & 27.9 & 22.7 & 22.5 & 27.8 & 26.0 & 22.6 & 24.2 & 21.1 \\
            MaskBlock & 23.8 & 21.2 & 20.8 & 23.0 & 23.6 & 22.4 & 22.1 & 20.5 \\
            US-MM & 27.6 & 23.8 & 23.7 & 26.4 & 26.8 & 24.1 & 25.8 & 21.8 \\
            
            BSR & 27.8 & 26.2 & 25.1 & 26.7 & 27.3 & 26.0 & 26.1 & 24.9 \\
            CWT & \textbf{30.9} & \textbf{28.7} & \textbf{27.9} & \textbf{30.8} & \textbf{29.3} & \textbf{27.8} & \textbf{29.5} & \textbf{28.7} \\
            \bottomrule
        \end{tabular}
    }
    \caption{Attack success rates (\%) of adversarial examples generated using various attack methods across eight classification models under \textbf{RS}. The defense model used is ResNet-50 with a noise level of 0.50.}
    \label{tab:RS}
\end{table}

\begin{figure*}[!ht]
\begin{center}
\vspace{-0.5em}
  \includegraphics[width=\textwidth]{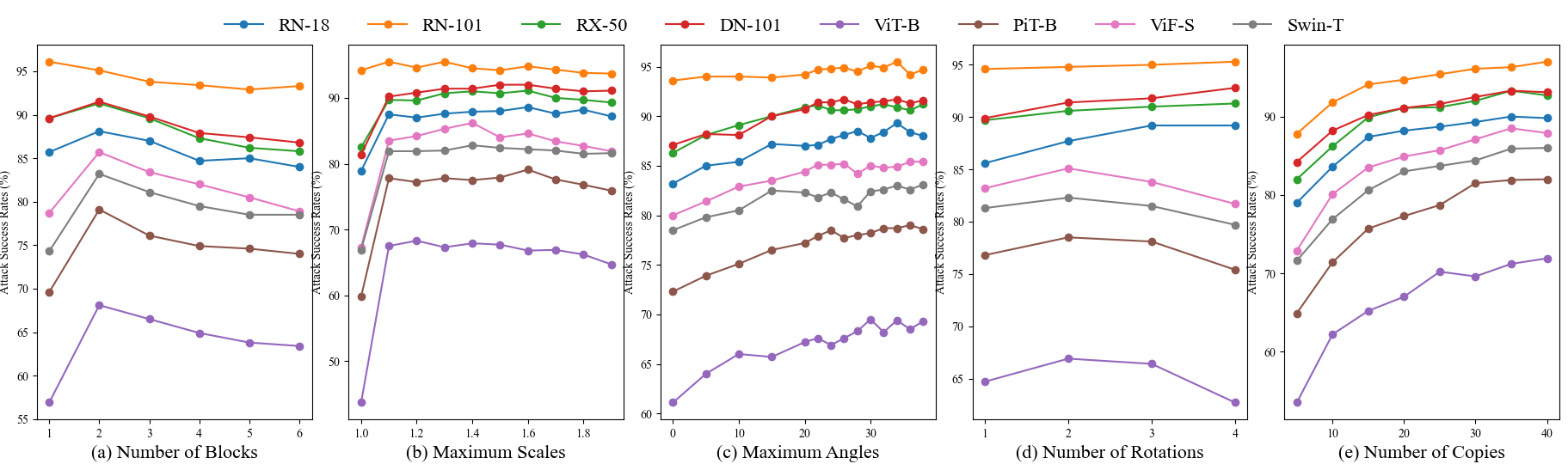} %插入图片，[]中设置图片大小，{}中是图片文件
\end{center}
% \vspace{-0.5em}
\caption{Attack success rates (\%) of various models on the adversarial examples generated by CWT with different numbers of blocks, maximum resize rates, maximum rotate angles, numbers of rotation blocks and numbers of transformed copies. The adversarial examples are crafted using the RN-101 model and tested on seven other models under the black-box setting.}
\label{fig:ablation}
% \vspace{-0.5em}
\end{figure*}

\paragraph{On the number of blocks $n$.}
The impact of block partitioning is illustrated in Figure~\ref{fig:ablation} (a). When $n = 1$, CWT applies global operations to the entire image, which fails to sufficiently disrupt the attention heatmaps. Increasing $n$ to $2$ significantly improves the attack success rate , as the $2 \times 2$ block division introduces localized transformations, encouraging the model to focus on diverse regions of the object.

When $n > 2$, performance begins to degrade due to increased information loss from operating on a larger number of blocks. However, in contrast to BSR, the performance does not drop sharply, as our selective rotation method stabilizes the results. Consequently, we set $n = 2$.

\paragraph{On the maximum scale factor $s_{\text{max}}$.}
As shown in Figure~\ref{fig:ablation} (b), the maximum scale factor significantly impacts the ASR. When $s_{\text{max}} > 1.4$, the ASR starts to decline gradually. This decline is due to the larger resizing requiring a random cropping operation, and as the image is enlarged, the cropped areas become larger, leading to information loss. Therefore, we set $s_{\text{max}} = 1.3$ as the optimal value for our experiments.

\paragraph{On the maximum rotation angle \( r \).}
As demonstrated in Figure~\ref{fig:ablation} (c), we observe that gradually increasing \( r \) improves ASR, but fluctuations occur when the angle exceeds 25°. Considering the experimental error margins, we interpret this fluctuation as insignificant, and thus we select \( r = 26^\circ \).

\paragraph{On the number of rotated blocks $k$.}
As shown in Figure~\ref{fig:ablation} (d), we find that increasing \( k \) generally leads to a decline in performance on most target models. This is due to the additional padding introduced by more rotation operations, which results in increased information loss. Therefore, we select \( k = 2 \) as the optimal value.

\paragraph{On the number of transformed copies $N$.}
As shown in Figure~\ref{fig:ablation} (e), the ASR steadily improves with increasing $N$, as it stabilizes gradient updates and introduces greater diversity. However, to maintain consistency with the BSR baseline and minimize computational overhead, we set $N = 20$ in our experiments.

\begin{figure}[!h]
    \centering
    % Uncomment and update the path to include the correct image file
    % \includegraphics[width=7.5cm]{figures/pos_attention_head.png}
    \includegraphics[width=8cm]{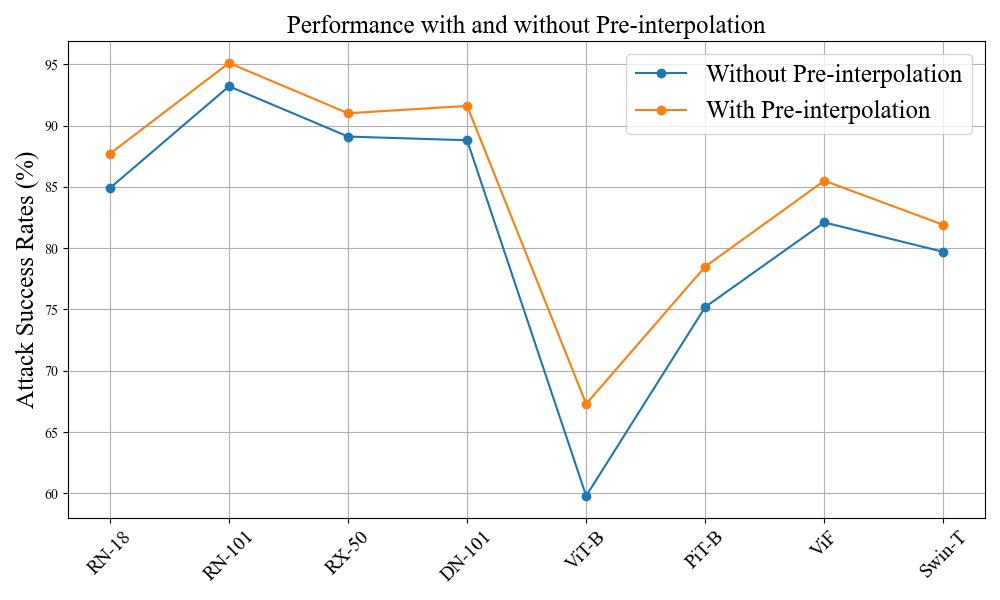}
    \caption{Attack success rates (\%) of various models on the adversarial examples generated by CWT with and without Pre-interpolation. The surrogate model is RN-101.}
    % \vspace{-0.2em}
    \label{fig:interpolation}
\end{figure}

\paragraph{On pre-interpolation.}
As shown in Figure~\ref{fig:interpolation}, we observe that pre-interpolation significantly improves the performance of the method. This demonstrates its ability to effectively remove redundant information, thereby enabling the model to focus on the relevant regions of the transformed image.

\section{Conclusion}

% 基于影响对抗样本的迁移性的主要障碍在于不同模型的注意力分布不一致这一发现，我们提出了CWT，它基于分块放大和低损旋转操作，能够有效的生成多样的图像，丰富单个分类模型的注意力分布，以消除不同模型在计算梯度损失时的差异，通过实验验证，我们的方法不仅在ASR上优于现有的SOTA，还在标准差上具有显著优势，这揭示了我们的方法的稳定性与泛化性，同时，我们的方法也为Input transformation-based 带来了新的视角。

Based on the observation that inconsistent attention distributions across models are a major barrier to adversarial example transferability, we propose CWT, a method utilizing interpolation  
 and selective rotation. CWT effectively generates diverse transformed images, enabling a single surrogate model to focus on different regions of the target object. Experimental results demonstrate that our approach not only surpasses existing SOTA methods in ASR but also achieves significantly lower standard deviation, highlighting its stability and generalization capabilities. Furthermore, CWT offers a novel perspective for input transformation-based adversarial attacks.
\clearpage
{
    \small
    \bibliographystyle{ieeenat_fullname}
    \bibliography{main}

\begin{thebibliography}{56}
\providecommand{\natexlab}[1]{#1}
\providecommand{\url}[1]{\texttt{#1}}
\expandafter\ifx\csname urlstyle\endcsname\relax
  \providecommand{\doi}[1]{doi: #1}\else
  \providecommand{\doi}{doi: \begingroup \urlstyle{rm}\Url}\fi

\bibitem[Chen et~al.(2023)Chen, Yin, Chen, Chen, and Liu]{chen2023adaptive}
Bin Chen, Jiali Yin, Shukai Chen, Bohao Chen, and Ximeng Liu.
\newblock An adaptive model ensemble adversarial attack for boosting adversarial transferability.
\newblock In \emph{Proceedings of the IEEE/CVF International Conference on Computer Vision}, pages 4489--4498, 2023.

\bibitem[Chen et~al.(2021)Chen, Xie, Niu, Liu, Wei, and Tian]{chen2021visformer}
Zhengsu Chen, Lingxi Xie, Jianwei Niu, Xuefeng Liu, Longhui Wei, and Qi Tian.
\newblock Visformer: The vision-friendly transformer.
\newblock In \emph{Proceedings of the IEEE/CVF international conference on computer vision}, pages 589--598, 2021.

\bibitem[Cohen et~al.(2019)Cohen, Rosenfeld, and Kolter]{cohen2019certified}
Jeremy Cohen, Elan Rosenfeld, and Zico Kolter.
\newblock Certified adversarial robustness via randomized smoothing.
\newblock In \emph{international conference on machine learning}, pages 1310--1320. PMLR, 2019.

\bibitem[Dong et~al.(2018)Dong, Liao, Pang, Su, Zhu, Hu, and Li]{dong2018boosting}
Yinpeng Dong, Fangzhou Liao, Tianyu Pang, Hang Su, Jun Zhu, Xiaolin Hu, and Jianguo Li.
\newblock Boosting adversarial attacks with momentum.
\newblock In \emph{Proceedings of the IEEE conference on computer vision and pattern recognition}, pages 9185--9193, 2018.

\bibitem[Dosovitskiy(2020)]{dosovitskiy2020image}
Alexey Dosovitskiy.
\newblock An image is worth 16x16 words: Transformers for image recognition at scale.
\newblock \emph{arXiv preprint arXiv:2010.11929}, 2020.

\bibitem[Eykholt et~al.(2018)Eykholt, Evtimov, Fernandes, Li, Rahmati, Xiao, Prakash, Kohno, and Song]{eykholt2018robust}
Kevin Eykholt, Ivan Evtimov, Earlence Fernandes, Bo Li, Amir Rahmati, Chaowei Xiao, Atul Prakash, Tadayoshi Kohno, and Dawn Song.
\newblock Robust physical-world attacks on deep learning visual classification.
\newblock In \emph{Proceedings of the IEEE conference on computer vision and pattern recognition}, pages 1625--1634, 2018.

\bibitem[Fan et~al.(2022)Fan, Chen, Liu, and Guo]{fan2022maskblock}
Mingyuan Fan, Cen Chen, Ximeng Liu, and Wenzhong Guo.
\newblock Maskblock: Transferable adversarial examples with bayes approach.
\newblock \emph{arXiv preprint arXiv:2208.06538}, 2022.

\bibitem[Gao et~al.(2020)Gao, Zhang, Song, Liu, and Shen]{gao2020patch}
Lianli Gao, Qilong Zhang, Jingkuan Song, Xianglong Liu, and Heng~Tao Shen.
\newblock Patch-wise attack for fooling deep neural network.
\newblock In \emph{Computer Vision--ECCV 2020: 16th European Conference, Glasgow, UK, August 23--28, 2020, Proceedings, Part XXVIII 16}, pages 307--322. Springer, 2020.

\bibitem[Geirhos et~al.(2018)Geirhos, Rubisch, Michaelis, Bethge, Wichmann, and Brendel]{geirhos2018imagenet}
Robert Geirhos, Patricia Rubisch, Claudio Michaelis, Matthias Bethge, Felix~A Wichmann, and Wieland Brendel.
\newblock Imagenet-trained cnns are biased towards texture; increasing shape bias improves accuracy and robustness.
\newblock In \emph{International conference on learning representations}, 2018.

\bibitem[Goodfellow et~al.(2014)Goodfellow, Shlens, and Szegedy]{goodfellow2014explaining}
Ian~J Goodfellow, Jonathon Shlens, and Christian Szegedy.
\newblock Explaining and harnessing adversarial examples.
\newblock \emph{arXiv preprint arXiv:1412.6572}, 2014.

\bibitem[Gowal et~al.(2019)Gowal, Dvijotham, Stanforth, Bunel, Qin, Uesato, Arandjelovic, Mann, and Kohli]{gowal2019scalable}
Sven Gowal, Krishnamurthy~Dj Dvijotham, Robert Stanforth, Rudy Bunel, Chongli Qin, Jonathan Uesato, Relja Arandjelovic, Timothy Mann, and Pushmeet Kohli.
\newblock Scalable verified training for provably robust image classification.
\newblock In \emph{Proceedings of the IEEE/CVF International Conference on Computer Vision}, pages 4842--4851, 2019.

\bibitem[He et~al.(2016)He, Zhang, Ren, and Sun]{he2016deep}
Kaiming He, Xiangyu Zhang, Shaoqing Ren, and Jian Sun.
\newblock Deep residual learning for image recognition.
\newblock In \emph{Proceedings of the IEEE conference on computer vision and pattern recognition}, pages 770--778, 2016.

\bibitem[Heo et~al.(2021)Heo, Yun, Han, Chun, Choe, and Oh]{heo2021rethinking}
Byeongho Heo, Sangdoo Yun, Dongyoon Han, Sanghyuk Chun, Junsuk Choe, and Seong~Joon Oh.
\newblock Rethinking spatial dimensions of vision transformers.
\newblock In \emph{Proceedings of the IEEE/CVF international conference on computer vision}, pages 11936--11945, 2021.

\bibitem[Hu et~al.(2018)Hu, Shen, and Sun]{hu2018squeeze}
Jie Hu, Li Shen, and Gang Sun.
\newblock Squeeze-and-excitation networks.
\newblock In \emph{Proceedings of the IEEE conference on computer vision and pattern recognition}, pages 7132--7141, 2018.

\bibitem[Huang et~al.(2017)Huang, Liu, Van Der~Maaten, and Weinberger]{huang2017densely}
Gao Huang, Zhuang Liu, Laurens Van Der~Maaten, and Kilian~Q Weinberger.
\newblock Densely connected convolutional networks.
\newblock In \emph{Proceedings of the IEEE conference on computer vision and pattern recognition}, pages 4700--4708, 2017.

\bibitem[Huh et~al.(2024)Huh, Cheung, Wang, and Isola]{huh2024platonic}
Minyoung Huh, Brian Cheung, Tongzhou Wang, and Phillip Isola.
\newblock The platonic representation hypothesis.
\newblock \emph{arXiv preprint arXiv:2405.07987}, 2024.

\bibitem[Kurakin et~al.(2018)Kurakin, Goodfellow, and Bengio]{kurakin2018adversarial}
Alexey Kurakin, Ian~J Goodfellow, and Samy Bengio.
\newblock Adversarial examples in the physical world.
\newblock In \emph{Artificial intelligence safety and security}, pages 99--112. Chapman and Hall/CRC, 2018.

\bibitem[Li et~al.(2023)Li, Guo, Zuo, and Chen]{li2023making}
Qizhang Li, Yiwen Guo, Wangmeng Zuo, and Hao Chen.
\newblock Making substitute models more bayesian can enhance transferability of adversarial examples.
\newblock \emph{arXiv preprint arXiv:2302.05086}, 2023.

\bibitem[Liao et~al.(2018)Liao, Liang, Dong, Pang, Hu, and Zhu]{liao2018defense}
Fangzhou Liao, Ming Liang, Yinpeng Dong, Tianyu Pang, Xiaolin Hu, and Jun Zhu.
\newblock Defense against adversarial attacks using high-level representation guided denoiser.
\newblock In \emph{Proceedings of the IEEE conference on computer vision and pattern recognition}, pages 1778--1787, 2018.

\bibitem[Lin et~al.(2019)Lin, Song, He, Wang, and Hopcroft]{lin2019nesterov}
Jiadong Lin, Chuanbiao Song, Kun He, Liwei Wang, and John~E Hopcroft.
\newblock Nesterov accelerated gradient and scale invariance for adversarial attacks.
\newblock \emph{arXiv preprint arXiv:1908.06281}, 2019.

\bibitem[Liu et~al.(2016)Liu, Chen, Liu, and Song]{liu2016delving}
Yanpei Liu, Xinyun Chen, Chang Liu, and Dawn Song.
\newblock Delving into transferable adversarial examples and black-box attacks.
\newblock \emph{arXiv preprint arXiv:1611.02770}, 2016.

\bibitem[Liu et~al.(2021)Liu, Lin, Cao, Hu, Wei, Zhang, Lin, and Guo]{liu2021swin}
Ze Liu, Yutong Lin, Yue Cao, Han Hu, Yixuan Wei, Zheng Zhang, Stephen Lin, and Baining Guo.
\newblock Swin transformer: Hierarchical vision transformer using shifted windows.
\newblock In \emph{Proceedings of the IEEE/CVF international conference on computer vision}, pages 10012--10022, 2021.

\bibitem[Long et~al.(2015)Long, Shelhamer, and Darrell]{long2015fully}
Jonathan Long, Evan Shelhamer, and Trevor Darrell.
\newblock Fully convolutional networks for semantic segmentation.
\newblock In \emph{Proceedings of the IEEE conference on computer vision and pattern recognition}, pages 3431--3440, 2015.

\bibitem[Ma et~al.(2025)Ma, Farahmand, Pan, Torr, and Gu]{ma2025improving}
Avery Ma, Amir-massoud Farahmand, Yangchen Pan, Philip Torr, and Jindong Gu.
\newblock Improving adversarial transferability via model alignment.
\newblock In \emph{European Conference on Computer Vision}, pages 74--92. Springer, 2025.

\bibitem[Madry(2017)]{madry2017towards}
Aleksander Madry.
\newblock Towards deep learning models resistant to adversarial attacks.
\newblock \emph{arXiv preprint arXiv:1706.06083}, 2017.

\bibitem[Naseer et~al.(2020)Naseer, Khan, Hayat, Khan, and Porikli]{naseer2020self}
Muzammal Naseer, Salman Khan, Munawar Hayat, Fahad~Shahbaz Khan, and Fatih Porikli.
\newblock A self-supervised approach for adversarial robustness.
\newblock In \emph{Proceedings of the IEEE/CVF Conference on Computer Vision and Pattern Recognition}, pages 262--271, 2020.

\bibitem[Nie et~al.(2022)Nie, Guo, Huang, Xiao, Vahdat, and Anandkumar]{nie2022diffusion}
Weili Nie, Brandon Guo, Yujia Huang, Chaowei Xiao, Arash Vahdat, and Anima Anandkumar.
\newblock Diffusion models for adversarial purification.
\newblock \emph{arXiv preprint arXiv:2205.07460}, 2022.

\bibitem[Proulx(2010)]{proulx2010size}
Michael~J Proulx.
\newblock Size matters: large objects capture attention in visual search.
\newblock \emph{PloS one}, 5\penalty0 (12):\penalty0 e15293, 2010.

\bibitem[Raghunathan et~al.(2018)Raghunathan, Steinhardt, and Liang]{raghunathan2018certified}
Aditi Raghunathan, Jacob Steinhardt, and Percy Liang.
\newblock Certified defenses against adversarial examples.
\newblock \emph{arXiv preprint arXiv:1801.09344}, 2018.

\bibitem[Redmon(2018)]{redmon2018yolov3}
Joseph Redmon.
\newblock Yolov3: An incremental improvement.
\newblock \emph{arXiv preprint arXiv:1804.02767}, 2018.

\bibitem[Ren et~al.(2016)Ren, He, Girshick, and Sun]{ren2016faster}
Shaoqing Ren, Kaiming He, Ross Girshick, and Jian Sun.
\newblock Faster r-cnn: Towards real-time object detection with region proposal networks.
\newblock \emph{IEEE transactions on pattern analysis and machine intelligence}, 39\penalty0 (6):\penalty0 1137--1149, 2016.

\bibitem[Ronneberger et~al.(2015)Ronneberger, Fischer, and Brox]{ronneberger2015u}
Olaf Ronneberger, Philipp Fischer, and Thomas Brox.
\newblock U-net: Convolutional networks for biomedical image segmentation.
\newblock In \emph{Medical image computing and computer-assisted intervention--MICCAI 2015: 18th international conference, Munich, Germany, October 5-9, 2015, proceedings, part III 18}, pages 234--241. Springer, 2015.

\bibitem[Salzmann et~al.(2021)]{salzmann2021learning}
Mathieu Salzmann et~al.
\newblock Learning transferable adversarial perturbations.
\newblock \emph{Advances in Neural Information Processing Systems}, 34:\penalty0 13950--13962, 2021.

\bibitem[Selvaraju et~al.(2016)Selvaraju, Das, Vedantam, Cogswell, Parikh, and Batra]{selvaraju2016grad}
Ramprasaath~R Selvaraju, Abhishek Das, Ramakrishna Vedantam, Michael Cogswell, Devi Parikh, and Dhruv Batra.
\newblock Grad-cam: Why did you say that?
\newblock \emph{arXiv preprint arXiv:1611.07450}, 2016.

\bibitem[Shafahi et~al.(2019)Shafahi, Najibi, Ghiasi, Xu, Dickerson, Studer, Davis, Taylor, and Goldstein]{shafahi2019adversarial}
Ali Shafahi, Mahyar Najibi, Mohammad~Amin Ghiasi, Zheng Xu, John Dickerson, Christoph Studer, Larry~S Davis, Gavin Taylor, and Tom Goldstein.
\newblock Adversarial training for free!
\newblock \emph{Advances in neural information processing systems}, 32, 2019.

\bibitem[Sharif et~al.(2016)Sharif, Bhagavatula, Bauer, and Reiter]{sharif2016accessorize}
Mahmood Sharif, Sruti Bhagavatula, Lujo Bauer, and Michael~K Reiter.
\newblock Accessorize to a crime: Real and stealthy attacks on state-of-the-art face recognition.
\newblock In \emph{Proceedings of the 2016 acm sigsac conference on computer and communications security}, pages 1528--1540, 2016.

\bibitem[Szegedy(2013)]{szegedy2013intriguing}
C Szegedy.
\newblock Intriguing properties of neural networks.
\newblock \emph{arXiv preprint arXiv:1312.6199}, 2013.

\bibitem[Tram{\`e}r et~al.(2017)Tram{\`e}r, Kurakin, Papernot, Goodfellow, Boneh, and McDaniel]{tramer2017ensemble}
Florian Tram{\`e}r, Alexey Kurakin, Nicolas Papernot, Ian Goodfellow, Dan Boneh, and Patrick McDaniel.
\newblock Ensemble adversarial training: Attacks and defenses.
\newblock \emph{arXiv preprint arXiv:1705.07204}, 2017.

\bibitem[Wang et~al.(2024)Wang, He, Wang, and Wang]{wang2024boosting}
Kunyu Wang, Xuanran He, Wenxuan Wang, and Xiaosen Wang.
\newblock Boosting adversarial transferability by block shuffle and rotation.
\newblock In \emph{Proceedings of the IEEE/CVF Conference on Computer Vision and Pattern Recognition}, pages 24336--24346, 2024.

\bibitem[Wang et~al.(2023{\natexlab{a}})Wang, Ying, Li, et~al.]{wang2023boost}
Tao Wang, Zijian Ying, Qianmu Li, et~al.
\newblock Boost adversarial transferability by uniform scale and mix mask method.
\newblock \emph{arXiv preprint arXiv:2311.12051}, 2023{\natexlab{a}}.

\bibitem[Wang and He(2021)]{wang2021enhancing}
Xiaosen Wang and Kun He.
\newblock Enhancing the transferability of adversarial attacks through variance tuning.
\newblock In \emph{Proceedings of the IEEE/CVF conference on computer vision and pattern recognition}, pages 1924--1933, 2021.

\bibitem[Wang et~al.(2021{\natexlab{a}})Wang, He, Wang, and He]{wang2021admix}
Xiaosen Wang, Xuanran He, Jingdong Wang, and Kun He.
\newblock Admix: Enhancing the transferability of adversarial attacks.
\newblock In \emph{Proceedings of the IEEE/CVF International Conference on Computer Vision}, pages 16158--16167, 2021{\natexlab{a}}.

\bibitem[Wang et~al.(2021{\natexlab{b}})Wang, Song, Wang, and He]{wang2021multi}
Xiaosen Wang, Chuanbiao Song, Liwei Wang, and Kun He.
\newblock Multi-stage optimization based adversarial training.
\newblock \emph{arXiv preprint arXiv:2106.15357}, 2021{\natexlab{b}}.

\bibitem[Wang et~al.(2023{\natexlab{b}})Wang, Zhang, and Zhang]{wang2023structure}
Xiaosen Wang, Zeliang Zhang, and Jianping Zhang.
\newblock Structure invariant transformation for better adversarial transferability.
\newblock In \emph{Proceedings of the IEEE/CVF International Conference on Computer Vision}, pages 4607--4619, 2023{\natexlab{b}}.

\bibitem[Wang et~al.(2023{\natexlab{c}})Wang, Pang, Du, Lin, Liu, and Yan]{wang2023better}
Zekai Wang, Tianyu Pang, Chao Du, Min Lin, Weiwei Liu, and Shuicheng Yan.
\newblock Better diffusion models further improve adversarial training.
\newblock In \emph{International Conference on Machine Learning}, pages 36246--36263. PMLR, 2023{\natexlab{c}}.

\bibitem[Wei and Zhao(2023)]{wei2023boosting}
Xingxing Wei and Shiji Zhao.
\newblock Boosting adversarial transferability with learnable patch-wise masks.
\newblock \emph{IEEE Transactions on Multimedia}, 2023.

\bibitem[Wei et~al.(2022)Wei, Chen, Goldblum, Wu, Goldstein, and Jiang]{wei2022towards}
Zhipeng Wei, Jingjing Chen, Micah Goldblum, Zuxuan Wu, Tom Goldstein, and Yu-Gang Jiang.
\newblock Towards transferable adversarial attacks on vision transformers.
\newblock In \emph{Proceedings of the AAAI Conference on Artificial Intelligence}, pages 2668--2676, 2022.

\bibitem[Wu et~al.(2020)Wu, Wang, Xia, Bailey, and Ma]{wu2020skip}
Dongxian Wu, Yisen Wang, Shu-Tao Xia, James Bailey, and Xingjun Ma.
\newblock Skip connections matter: On the transferability of adversarial examples generated with resnets.
\newblock \emph{arXiv preprint arXiv:2002.05990}, 2020.

\bibitem[Xie et~al.(2019{\natexlab{a}})Xie, Wu, Maaten, Yuille, and He]{xie2019feature}
Cihang Xie, Yuxin Wu, Laurens van~der Maaten, Alan~L Yuille, and Kaiming He.
\newblock Feature denoising for improving adversarial robustness.
\newblock In \emph{Proceedings of the IEEE/CVF conference on computer vision and pattern recognition}, pages 501--509, 2019{\natexlab{a}}.

\bibitem[Xie et~al.(2019{\natexlab{b}})Xie, Zhang, Zhou, Bai, Wang, Ren, and Yuille]{xie2019improving}
Cihang Xie, Zhishuai Zhang, Yuyin Zhou, Song Bai, Jianyu Wang, Zhou Ren, and Alan~L Yuille.
\newblock Improving transferability of adversarial examples with input diversity.
\newblock In \emph{Proceedings of the IEEE/CVF conference on computer vision and pattern recognition}, pages 2730--2739, 2019{\natexlab{b}}.

\bibitem[Xie et~al.(2017)Xie, Girshick, Doll{\'a}r, Tu, and He]{xie2017aggregated}
Saining Xie, Ross Girshick, Piotr Doll{\'a}r, Zhuowen Tu, and Kaiming He.
\newblock Aggregated residual transformations for deep neural networks.
\newblock In \emph{Proceedings of the IEEE conference on computer vision and pattern recognition}, pages 1492--1500, 2017.

\bibitem[Xiong et~al.(2022)Xiong, Lin, Zhang, Hopcroft, and He]{xiong2022stochastic}
Yifeng Xiong, Jiadong Lin, Min Zhang, John~E Hopcroft, and Kun He.
\newblock Stochastic variance reduced ensemble adversarial attack for boosting the adversarial transferability.
\newblock In \emph{Proceedings of the IEEE/CVF conference on computer vision and pattern recognition}, pages 14983--14992, 2022.

\bibitem[Yuan et~al.(2022)Yuan, Zhang, Gao, Cheng, and Song]{yuan2022natural}
Shengming Yuan, Qilong Zhang, Lianli Gao, Yaya Cheng, and Jingkuan Song.
\newblock Natural color fool: Towards boosting black-box unrestricted attacks.
\newblock \emph{Advances in Neural Information Processing Systems}, 35:\penalty0 7546--7560, 2022.

\bibitem[Zhang et~al.(2023)Zhang, Huang, Wang, Li, Wu, Wang, Su, and Lyu]{zhang2023improving}
Jianping Zhang, Jen-tse Huang, Wenxuan Wang, Yichen Li, Weibin Wu, Xiaosen Wang, Yuxin Su, and Michael~R Lyu.
\newblock Improving the transferability of adversarial samples by path-augmented method.
\newblock In \emph{Proceedings of the IEEE/CVF Conference on Computer Vision and Pattern Recognition}, pages 8173--8182, 2023.

\bibitem[Zhang et~al.(2025)Zhang, Khayatkhoei, Chhikara, and Ilievski]{zhang2025mllms}
Jiarui Zhang, Mahyar Khayatkhoei, Prateek Chhikara, and Filip Ilievski.
\newblock Mllms know where to look: Training-free perception of small visual details with multimodal llms.
\newblock \emph{arXiv preprint arXiv:2502.17422}, 2025.

\bibitem[Zhu et~al.(2024)Zhu, Zhang, Liang, Liu, and Xu]{zhu2024learning}
Rongyi Zhu, Zeliang Zhang, Susan Liang, Zhuo Liu, and Chenliang Xu.
\newblock Learning to transform dynamically for better adversarial transferability.
\newblock In \emph{Proceedings of the IEEE/CVF Conference on Computer Vision and Pattern Recognition}, pages 24273--24283, 2024.

\end{thebibliography}
}
% \clearpage
% \setcounter{page}{1}
% \maketitlesupplementary

\clearpage
\setcounter{page}{1}
\maketitlesupplementary

\appendix
\setcounter{table}{0}
\renewcommand{\thetable}{A\arabic{table}}
\setcounter{figure}{0}
\renewcommand{\thefigure}{A\arabic{figure}}

% \section{Rationale}
% \label{sec:rationale}
% % 
% Having the supplementary compiled together with the main paper means that:
% % 
% \begin{itemize}
% \item The supplementary can back-reference sections of the main paper, for example, we can refer to \cref{sec:intro};
% \item The main paper can forward reference sub-sections within the supplementary explicitly (e.g. referring to a particular experiment); 
% \item When submitted to arXiv, the supplementary will already included at the end of the paper.
% \end{itemize}
% % 
% To split the supplementary pages from the main paper, you can use \href{https://support.apple.com/en-ca/guide/preview/prvw11793/mac#:~:text=Delete%20a%20page%20from%20a,or%20choose%20Edit%20%3E%20Delete).}{Preview (on macOS)}, \href{https://www.adobe.com/acrobat/how-to/delete-pages-from-pdf.html#:~:text=Choose%20%E2%80%9CTools%E2%80%9D%20%3E%20%E2%80%9COrganize,or%20pages%20from%20the%20file.}{Adobe Acrobat} (on all OSs), as well as \href{https://superuser.com/questions/517986/is-it-possible-to-delete-some-pages-of-a-pdf-document}{command line tools}.

\section{Extended Evaluations on Defense Methods}
\label{Extended-Evaluations}
% In this section, we present the performance of our method on 6个adversarially trained models  i.e.
% (Inc-v3\(_{ens}\),Inc-v3\(_{ens3}\),Inc-v3\(_{ens4}\)},IncRes-v2\(_{ens}\)},Res50$_{SIN}$},Res50$_{IN}$)~\cite{tramer2017ensemble}以及NRP, as shown in Tables ~\ref{tab:adversarially trained models.}through ~\ref{tab:NRPVit}. 对于Table ~\ref{tab:adversarially trained models.}, 我们测试不同攻击方法在不同代理模型下生成的对抗样本对于经过对抗训练的模型的攻击效果。值得注意的是，这些模型无法对于ground true的图像进行百分之百的分类，但是除了res50_sin分类错误的比率为 21.5\%外,其余错误的比率都不超过5\%。而表中的结果显示，我们的方法相比与之前的SOTA平均提高超过5\%. 对于Tables ~\ref{tab:NRPCnn} and ~\ref{tab:NRPVit}, we apply NRP to process adversarial images generated by different defense methods, and subsequently evaluate their Attack Success Rate (ASR) using eight models. The results demonstrate that our method, CWT, achieves superior performance in terms of both ASR and standard deviation (std) when applied to CNN-based and transformer-based surrogate models. Notably, while maintaining a reduction in standard deviation, our approach consistently improves the ASR by an average of 5\% over the previous state-of-the-art methods.

In this section, we evaluate the performance of our proposed method on five adversarially trained models, namely Inc-v3\(_{ens}\), Inc-v3\(_{ens3}\), Inc-v3\(_{ens4}\), IncRes-v2\(_{ens}\) and Res50\(_{in}\) ~\cite{tramer2017ensemble,geirhos2018imagenet}, as well as the NRP~\cite{naseer2020self} method. The results are summarized in Tables ~\ref{tab:adversarially_trained_models} through ~\ref{tab:NRPVit}.

Specifically, in Table ~\ref{tab:adversarially_trained_models}, we analyze the attack effectiveness of adversarial samples generated using different attack methods on various surrogate models against adversarially trained models. It is worth noting that these models do not achieve 100\% classification accuracy on clean images. However, the misclassification rates of the models are all below 5\%. The results indicate that our method surpasses previous SOTA approaches by an average improvement of 5.3\%.

For Tables~\ref{tab:NRPCnn} and ~\ref{tab:NRPVit}, we apply NRP to process adversarial images generated by different defense methods and ASR on eight models. The results demonstrate that our method, CWT, outperforms existing techniques in terms of both ASR and standard deviation for CNN-based and Transformer-based surrogate models. Notably, while maintaining a reduced standard deviation, our approach consistently improves the ASR by an average of more than 5\% over prior SOTA methods.

\begin{table*}[h!]
    \centering
    \scriptsize
    \setlength{\tabcolsep}{3pt}
    \renewcommand{\arraystretch}{0.7}
    \resizebox{\textwidth}{!}{
    \begin{tabular}{c c c c c c c c c c c c}
        \toprule
        \textbf{Model} & \textbf{Attack} & \textbf{RN-18 ($\uparrow$)} & \textbf{RN-101 ($\uparrow$)} & \textbf{RX-50 ($\uparrow$)} & \textbf{DN-121 ($\uparrow$)} & \textbf{ViT-B ($\uparrow$)} & \textbf{PiT-B ($\uparrow$)} & \textbf{ViF-S ($\uparrow$)} & \textbf{Swin-T ($\uparrow$)} & \textbf{Mean ($\uparrow$)} & \textbf{Std. Dev. ($\downarrow$)}\\
        \midrule
        % Block 1: Inc-v3_ens (取数据的第1个数值)
        \multirow{7}{*}{\textbf{Inc-v3\(_{ens}\)}} 
            & DIM       & 64.2   & 50.7   & 46.2   & 68.0   & 47.8   & 48.8   & 61.4   & 52.4   & 54.9   & 7.8 \\
            & SIM       & 60.1   & 45.2   & 41.2   & 67.2   & 49.0   & 44.1   & 54.8   & 28.6   & 48.8   & 11.2 \\
            & Admix     & 71.6   & 58.9   & 50.6   & 77.9   & 51.4   & 43.8   & 59.5   & 31.1   & 55.6   & 14.0 \\
            & MaskBlock & 44.4   & 36.3   & 31.1   & 51.0   & 42.5   & 39.2   & 41.8   & 25.9   & 39.0   & 7.4 \\
            & US-MM     & 68.7   & 60.1   & 52.8   & 74.4   & 53.5   & 47.2   & 64.8   & 30.8   & 56.5   & 12.9 \\
            & BSR       & 88.9   & 79.1   & 71.5   & 89.8   & 72.7   & 75.0   & 81.4   & 75.8   & 79.3   & 6.5 \\
            & CWT       & \textbf{92.7}   & \textbf{84.6}   & \textbf{79.0}   & \textbf{92.6}   & \textbf{74.0}   & \textbf{82.1}   & \textbf{86.1}   & \textbf{84.0}   & \textbf{84.4}   & \textbf{5.9} \\
        \midrule
        % Block 2: Inc-v3_ens3 (取数据的第2个数值)
        \multirow{7}{*}{\textbf{Inc-v3\(_{ens3}\)}} 
            & DIM       & 60.2   & 48.3   & 43.7   & 65.7   & 44.7   & 47.1   & 58.4   & 49.8   & 52.2   & 7.6 \\
            & SIM       & 57.1   & 43.7   & 36.4   & 63.8   & 46.7   & 41.2   & 51.7   & 26.1   & 45.8   & 11.1 \\
            & Admix     & 67.9   & 56.4   & 49.7   & 75.1   & 49.5   & 42.1   & 59.2   & 29.3   & 53.6   & 13.5 \\
            & MaskBlock & 43.4   & 36.2   & 28.5   & 51.1   & 36.5   & 38.4   & 40.9   & 24.8   & 37.5   & 7.7 \\
            & US-MM     & 65.0   & 60.1   & 51.2   & 72.3   & 50.1   & 45.2   & 61.9   & 29.8   & 54.5   & 12.5 \\
            & BSR       & 85.7   & 77.7   & 70.5   & 87.1   & 71.0   & 71.6   & 78.5   & 73.6   & 77.0   & 6.1 \\
            & CWT       & \textbf{91.4}   & \textbf{82.5}   & \textbf{77.5}   & \textbf{90.8}   & \textbf{72.5}   & \textbf{79.4}   & \textbf{85.5}   & \textbf{81.6}   & \textbf{82.7}   & \textbf{6.0} \\
        \midrule
        % Block 3: Inc-v3_ens4 (取数据的第3个数值)
        \multirow{7}{*}{\textbf{Inc-v3\(_{ens4}\)}} 
            & DIM       & 61.5   & 45.2   & 39.1   & 65.2   & 43.9   & 46.0   & 54.8   & 48.7   & 50.6   & 8.5 \\
            & SIM       & 52.7   & 39.8   & 34.4   & 62.8   & 44.5   & 38.2   & 48.6   & 25.8   & 43.4   & 10.7 \\
            & Admix     & 64.3   & 53.2   & 45.7   & 72.9   & 47.5   & 40.0   & 54.6   & 28.0   & 50.8   & 13.1 \\
            & MaskBlock & 39.1   & 32.3   & 27.8   & 47.0   & 36.8   & 33.7   & 39.3   & 24.9   & 35.1   & 6.6 \\
            & US-MM     & 61.6   & 56.8   & 46.8   & 71.0   & 49.7   & 43.5   & 58.5   & 27.5   & 51.9   & 12.4 \\
            & BSR       & 85.1   & 73.6   & 66.8   & 86.4   & 69.2   & 70.5   & 77.2   & 72.9   & 75.2   & 6.8 \\
            & CWT       & \textbf{90.9}   & \textbf{82.0}   & \textbf{75.4}   & \textbf{90.8}   & \textbf{72.1}   & \textbf{79.1}   & \textbf{81.7}   & \textbf{79.9}   & \textbf{81.5}   & \textbf{6.2} \\
        \midrule
        % Block 4: IncRes-v2_ens (取数据的第4个数值)
        \multirow{7}{*}{\textbf{IncRes-v2\(_{ens}\)}} 
            & DIM       & 48.0   & 43.9   & 34.8   & 55.1   & 41.5   & 39.8   & 50.2   & 39.8   & 44.1   & 6.2 \\
            & SIM       & 42.1   & 36.7   & 30.3   & 50.6   & 39.7   & 33.6   & 44.6   & 19.2   & 37.1   & 9.0 \\
            & Admix     & 50.6   & 50.1   & 41.6   & 61.8   & 44.9   & 34.2   & 50.6   & 21.0   & 44.4   & 11.6 \\
            & MaskBlock & 30.1   & 28.4   & 23.3   & 37.7   & 33.6   & 29.2   & 30.0   & 17.5   & 28.7   & 5.7 \\
            & US-MM     & 49.3   & 51.7   & 44.0   & 59.1   & 44.4   & 38.4   & 53.2   & 21.0   & 45.1   & 10.9 \\
            & BSR       & 75.0   & 71.4   & 63.9   & 77.8   & 67.4   & 66.1   & 69.5   & 64.0   & 69.4   & 4.8 \\
            & CWT       & \textbf{81.8}   & \textbf{79.6}   & \textbf{73.5}   & \textbf{84.6}   & \textbf{69.6}   & \textbf{73.8}   & \textbf{76.9}   & \textbf{73.9}   & \textbf{76.7}   & \textbf{4.7} \\
        \midrule
        % Block 4: IncRes-v2_{ens}
        \multirow{7}{*}{\textbf{Res50\(_{in}\)}} 
            & DIM       & 83.8   & 54.1   & 51.1   & 79.6   & 49.7   & 52.9   & 61.8   & 57.6   & 61.3   & 12.3 \\
            & SIM       & 81.6   & 53.9   & 49.3   & 79.1   & 54.9   & 46.8   & 57.6   & 36.2   & 57.4   & 14.6 \\
            & Admix     & 90.1   & 67.1   & 61.3   & 87.7   & 58.2   & 49.2   & 64.6   & 40.3   & 64.8   & 16.1 \\
            & MaskBlock & 72.4   & 47.1   & 39.8   & 70.5   & 48.3   & 44.7   & 47.6   & 33.3   & 50.5   & 13.0 \\
            & US-MM     & 88.2   & 73.1   & 66.8   & 87.7   & 60.7   & 52.6   & 70.7   & 41.4   & 67.7   & 15.1 \\
            & BSR       & 97.1   & 83.2   & 78.8   & 95.3   & \textbf{75.1}   & 77.0   & 82.3   & 82.3   & 83.9   & 7.6 \\
            & CWT       & \textbf{98.1}   & \textbf{85.7}   & \textbf{82.2}   & \textbf{95.3}   & 74.6   & \textbf{80.5}   & \textbf{85.7}   & \textbf{85.9}   & \textbf{86.0}   & \textbf{7.1} \\
        \bottomrule
    \end{tabular}
    }
    \caption{Attack success rates (\%) of adversarial examples generated using various attack methods across eight classification models under \textbf{adversarially trained models}.}
\label{tab:adversarially_trained_models}
\end{table*}

\begin{table*}[h!]
    \centering
    \scriptsize
    \setlength{\tabcolsep}{3pt}
    \renewcommand{\arraystretch}{0.7}
    \resizebox{\textwidth}{!}{
        \begin{tabular}{c c c c c c c c c c c c}
            \toprule
            \textbf{Model} & \textbf{Attack} & \textbf{RN-18 ($\uparrow$)} & \textbf{RN-101 ($\uparrow$)} & \textbf{RX-50 ($\uparrow$)} & \textbf{DN-121 ($\uparrow$)} & \textbf{ViT-B ($\uparrow$)} & \textbf{PiT-B ($\uparrow$)} & \textbf{ViF-S ($\uparrow$)} & \textbf{Swin-T ($\uparrow$)} & \textbf{Mean ($\uparrow$)} & \textbf{Std. Dev. ($\downarrow$)}\\
            \midrule
            \multirow{7}{*}{\textbf{RN-18}} 
              & DIM       & 98.9* & 39.5           & 40.7           & 64.5           & 19.5           & 25.8           & 30.4           & 34.7           & 44.3          & 25.8          \\
 & SIM       & 99.5* & 40.6           & 44.7           & 67.5           & 17.2           & 24.9           & 31.9           & 37.3           & 45.5          & 26.5          \\
 & Admix     & \textbf{99.9*} & 47.8           & 51.9           & 77.7           & 19.9           & 28.4           & 35.8           & 41.5           & 50.4          & 26.5          \\
 & MaskBlock & 98.0*   & 32             & 35.5           & 57.5           & 13.6           & 19             & 25             & 28.1           & 38.6          & 27.4          \\
& US-MM & 99.4* & 46.6 & 51.3 & 76.2 & 21.5 & 29.3 & 39.5 & 42.6 & 50.8 & 25.5 \\

 & BSR   & 98.8* & 53.4 & 60.0 & 81.8 & 25.9 & 37.7 & 46.7 & 50.0  &56.8&23.5        \\
 & CWT  & 99.4* & \textbf{58.6} & \textbf{64.9 }& \textbf{87.4} & \textbf{33.5 }& \textbf{43.7} & \textbf{52.4} & 5\textbf{5.2}&\textbf{61.9}&\textbf{21.9} \\
          \midrule
            \multirow{7}{*}{\textbf{RN-101}} 
             & DIM       & 54.4           & 56.8*          & 38.7           & 49.5           & 19.1           & 25             & 28.5           & 29.4           & 37.7          & 14.4          \\
 & SIM       & 55.7           & 62.6*          & 40.8           & 51             & 17.1           & 25.7           & 28.3           & 30.6           & 39.0          & 16.1          \\
 & Admix     & 60.9           & 75.7* & 52.9           & 58.9           & 22.6           & 33.2           & 37.9           & 39.8           & 47.7          & 17.4          \\
 & MaskBlock & 53.3           & 54.6*          & 35.6           & 47.3           & 13             & 19.2           & 23             & 24.9           & 33.9          & 16.2          \\
& US-MM & 99.4* & 46.6 & 51.3 & 76.2 & 21.5 & 29.3 & 39.5 & 42.6 & 50.8 & 25.5 \\

 & BSR   & 71.9 & 73.4* & 63.6 & 70.8 & 32.2 & 45.0 & 51.8 & 50.1  &57.7&14.9      \\
 & CWT & \textbf{74.7} & \textbf{80.9*} & \textbf{68.9} & \textbf{75.3} & \textbf{40.4} & \textbf{52.4} & \textbf{58.2} & \textbf{57.3} &\textbf{63.5}&\textbf{13.7}\\
            \midrule
            \multirow{7}{*}{\textbf{RX-50}} 
          & DIM       & 54             & 32.8           & 57.4*          & 46.8           & 16             & 23             & 26.8           & 27.2           & 35.5          & 15.3          \\
 & SIM       & 55.9           & 35.5           & 63.3*          & 48.3           & 14.9           & 22.6           & 25.5           & 28.1           & 36.8          & 17.3          \\
 & Admix     & 60.1           & 45.4           & 76.2* & 56.6           & 19.8           & 29.7           & 34.1           & 35.9           & 44.7          & 18.5          \\
 & MaskBlock & 53.4           & 29.3           & 55.2*          & 41.9           & 11.9           & 18.4           & 22.6           & 23.7           & 32.1          & 16.2          \\
 & US-MM & 68.9 & 49.0 & 78.8* & 63.0 & 31.3 & 37.1 & 39.9 & 48.6 & 48.7 & 20.0 \\

  & BSR   & 68.8 & 54.1 & 74.3* & 67.2 & 28.3 & 40.8 & 46.3 & 46.4  &53.3&15.8        \\
 & CWT  &  \textbf{70.8} & \textbf{59.8} & \textbf{80.4*} & \textbf{71.5} &\textbf{ 34.3} & \textbf{47.0} & \textbf{53.7} & \textbf{52.6 } &\textbf{58.8}&\textbf{15.0}\\
            \midrule
            \multirow{7}{*}{\textbf{DN-121}} 
               & DIM       & 68.4           & 41.2           & 45.3           & 97*            & 20.2           & 26.8           & 33.9           & 35.7           & 46.1          & 25.1          \\
 & SIM       & 74.8           & 45.5           & 49.6           & 98.6*          & 20.9           & 30.1           & 38.4           & 40.6           & 49.8          & 25.2          \\
 & Admix     & 81.3  & 55.9  & 57.3           & \textbf{99.3*} & 28             & 35             & 43.9           & 48.1           & 56.1          & 23.8          \\
 & MaskBlock & 67.5           & 35.3           & 39.1           & 95.1*          & 15             & 22.7           & 29.3           & 31.7           & 42.0          & 26.4          \\
 & US-MM & 82.0 & 54.4 & 58.6 & 98.7* & 27.5 & 35.2 & 45.8 & 47.8 & 56.4 & 23.8 \\

& BSR   & \textbf{83.8} & 56.3 & 61.0 & 96.6* & 29.9 & 38.6 & 48.5 & 48.7   &57.9&22.4      \\
 & CWT  & 83.1 & \textbf{60.9} & \textbf{64.7} & 98.6*& \textbf{35.1} &\textbf{ 46.2} & \textbf{52.9} & \textbf{54.8 }&\textbf{62.0}&\textbf{20.4}\\
            \bottomrule
        \end{tabular}
    }
        \caption{Attack success rates (\%) of adversarial examples generated using various attack methods across eight classification models under \textbf{NRP}. The surrogate models are \textbf{CNN-based.} * indicates the surrogate model.}
    \label{tab:NRPCnn}
\end{table*}

\begin{table*}[h!]
    \centering
    \scriptsize
    \setlength{\tabcolsep}{3pt}
    \renewcommand{\arraystretch}{0.7}
    \resizebox{\textwidth}{!}{
        \begin{tabular}{c c c c c c c c c c c c}
            \toprule
            \textbf{Model} & \textbf{Attack} & \textbf{RN-18 ($\uparrow$)} & \textbf{RN-101 ($\uparrow$)} & \textbf{RX-50 ($\uparrow$)} & \textbf{DN-121 ($\uparrow$)} & \textbf{ViT-B ($\uparrow$)} & \textbf{PiT-B ($\uparrow$)} & \textbf{ViF-S ($\uparrow$)} & \textbf{Swin-T ($\uparrow$)} & \textbf{Mean ($\uparrow$)} & \textbf{Std. Dev. ($\downarrow$)}\\
            \midrule
            \multirow{7}{*}{\textbf{ViT-B}}
               & DIM       & 52.5           & 34.3           & 34.1           & 47.9           & 70.9*          & 34.4           & 35.4           & 39             & 43.6          & 13.1          \\
 & SIM       & 57.6           & 36.1           & 39.2           & 52.5           & 85.3*          & 38.4           & 40.5           & 47             & 49.6          & 16.3          \\
 & Admix     & 57.9           & 37.9           & 41.4           & 54.2           & \textbf{86.9*} & 40.7           & 41.8           & 51.3           & 51.5          & 16.0          \\
 & MaskBlock & 56.8           & 30.3           & 32.7           & 47.2           & 81.6*          & 30.1           & 32.6           & 40.6           & 44.0          & 17.9          \\
& US-MM & 61.1 & 40.0 & 42.8 & 58.6 & 84.6* & 42.5 & 44.7 & 55.0 & 53.7 & 15.5 \\

& BSR  & 66.0 & 49.9 & 51.9 & 62.8 & 69.1* & 55.3 & 55.1 & 56.8  &58.4&\textbf{6.9}         \\
 & CWT  &  \textbf{66.7} & \textbf{52.1} & \textbf{53.8} & \textbf{64.1} & 75.9*& \textbf{59.1} & \textbf{57.7} & \textbf{58.8} &\textbf{61.0}&7.7     \\
            \midrule
            \multirow{7}{*}{\textbf{PiT-B}} 
              & DIM       & 56             & 32.2           & 36.1           & 48.1           & 27.2           & 67.8*          & 37.2           & 38.1           & 42.8          & 13.5          \\
 & SIM       & 53.1           & 32             & 34.5           & 45.8           & 24.7           & 73.5*          & 35.9           & 38.1           & 42.2          & 15.3          \\
 & Admix     & 54.8           & 34.7           & 36             & 47             & 25.8           & 73.8*          & 39.8           & 42.8           & 44.3          & 14.7          \\
 & MaskBlock & 56.8           & 29.7           & 34.2           & 44.5           & 23.1           & 76.5*          & 34.8           & 37.1           & 42.1          & 17.2          \\
 & US-MM & 68.4 & 37.4 & 39.2 & 52.2 & 29.4 & 74.4* & 43.6 & 46.0 & 47.6 & 14.0 \\

 & BSR   & 68.1 & 50.2 & 54.9 & 65.1 & 44.1 & 79.7*& 60.0 & 61.9  &60.5&11.1        \\
 & CWT  & \textbf{70.6} & \textbf{56.7} & \textbf{59.9} & \textbf{69.3} & \textbf{52.7} & \textbf{85.1*}& \textbf{65.8} & \textbf{67.3} &\textbf{65.9}&\textbf{10.0} \\
            \midrule
            \multirow{7}{*}{\textbf{ViF-S}} 
              & DIM       & 59.3           & 37.8           & 41.5           & 53.6           & 28.6           & 41.1           & 76.7*          & 44.9           & 47.9          & 14.9          \\
 & SIM       & 60.7           & 38.2           & 40.3           & 53             & 27.6           & 41.2           & 80.8*          & 47.8           & 48.7          & 16.4          \\
 & Admix     & 61.8           & 42.5           & 46.3           & 59.3           & 32             & 47.2           & 85.6*          & 54.1           & 53.6          & 16.1          \\
 & MaskBlock & 57.9           & 32.8           & 33.8           & 47.6           & 20.1           & 30.8           & 76.3*          & 38.5           & 42.2          & 17.8          \\
& US-MM & 68.8 & 46.5 & 52.7 & 64.1 & 35.3 & 50.9 & 86.4* & 58.7 & 57.9 & 15.5 \\

 & BSR   & 71.7 & 52.1 & 56.8 & 69.2 & 40.5 & 59.3 & 81.6* & 61.4&61.6&12.7       \\
 & CWT  & \textbf{73.4} & \textbf{59.0} & \textbf{63.0} & \textbf{73.1 }& \textbf{50.8}& \textbf{65.9} & \textbf{87.9*}& \textbf{68.9 } &\textbf{67.8}&\textbf{11.1}\\
            \midrule
            \multirow{7}{*}{\textbf{Swin-T}} 
              & DIM       & 54.6           & 31.5           & 33.2           & 46.1           & 24.6           & 33.5           & 37.4           & 74.5*          & 41.9          & 16.1          \\
 & SIM       & 45.7           & 25.4           & 25.9           & 37.4           & 14             & 22             & 24.5           & 70.4*          & 33.2          & 17.9          \\
 & Admix     & 50.9           & 25             & 26.7           & 40.5           & 15.2           & 23.8           & 27.7           & 75.2*          & 35.6          & 19.4          \\
 & MaskBlock & 50.1           & 24.1           & 24.9           & 38.3           & 13.5           & 20.8           & 24.9           & 69.1*          & 33.2          & 18.4          \\
& US-MM & 52.1 & 26.1 & 29.6 & 43.1 & 16.2 & 23.2 & 28.3 & 74.3* & 36.6 & 19.0 \\

 & BSR   & 68.8 & 48.6 & 51.5 & 64.6 & 36.7 & 54.7 & 57.0 & 76.9* &57.4&12.6        \\
 & CWT  & \textbf{70.9} & \textbf{54.6}&  \textbf{57.7} & \textbf{71.5} & \textbf{4}\textbf{5.1}& \textbf{62.4} & \textbf{62.6} & \textbf{83.1*}&\textbf{63.5}&\textbf{11.7}
\\
            \bottomrule
        \end{tabular}
    }
    \caption{Attack success rates (\%) of adversarial examples generated using various attack methods across eight classification models under \textbf{NRP}. The surrogate models are \textbf{Transformer-based}. * indicates the surrogate model.}
    \label{tab:NRPVit}
\end{table*}

\section{More Analysis on Interpolation}
\label{appedix: More Analysis on Interpolation}

\begin{figure*}[h]
\begin{center}
% \vspace{-1.2em}
  \includegraphics[width=\textwidth]{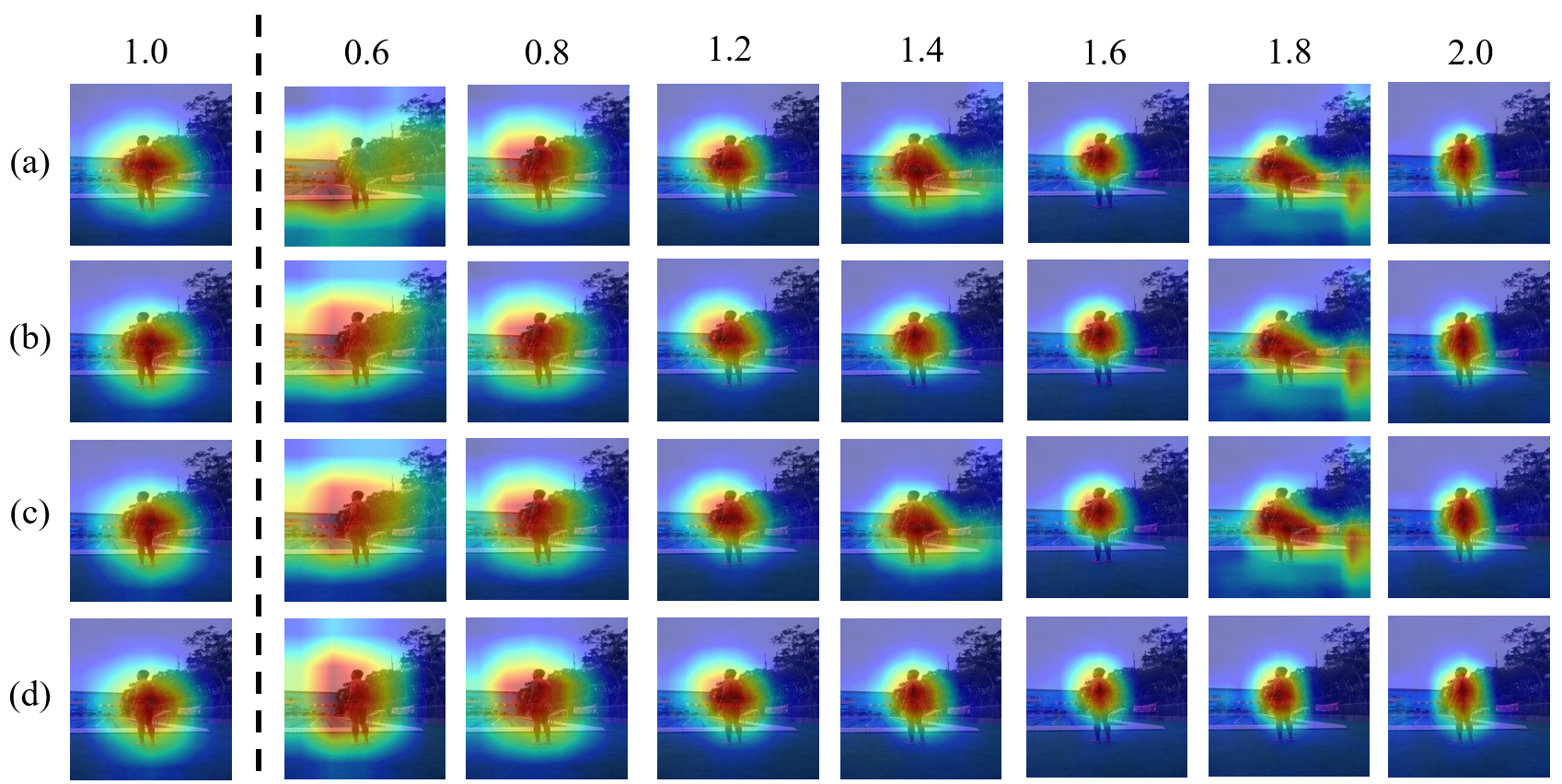} %插入图片，[]中设置图片大小，{}中是图片文件
\end{center}
% \vspace{-1.2em}
\caption{\textbf{Heatmaps of Different Interpolation Rates Using ResNet-18}. (a) Area interpolation; (b) Bicubic interpolation; (c) Bilinear interpolation; (d) Nearest-neighbor interpolation.}
\label{fig:interpolation-rate}
% \vspace{-1.5em}
\end{figure*}

To better understand the role of interpolation in affecting model attention distributions, we conducted experiments using various interpolation methods (e.g., bilinear, bicubic, nearest neighbor, and area interpolation) under different scaling factors. As shown in Figure~~\ref{fig:interpolation-rate}, scaling factors below 1.0 tend to disperse attention across irrelevant regions, reducing focus on critical object areas. Conversely, scaling factors significantly above 1.0 excessively concentrate attention on limited regions. Moderate scaling factors, typically ranging from 1.0 to 1.8, yield the most balanced attention distributions, effectively redistributing focus across diverse object regions. For example, as depicted in Figure~~\ref{fig:interpolation-rate} (a), a scaling factor of 1.0 primarily directs attention to a person’s back, while a factor of 1.4 shifts focus towards the hips, and 1.8 further moves attention to the left shoulder. 

In our method, we strategically apply a two-step process to achieve balanced attention redistribution. First, we perform a shrinking operation on each block to disperse attention and eliminate redundant information. Next, we enlarge the block, refocusing the previously dispersed attention on critical regions of the object. This two-step process effectively balances attention distribution and enhances the model’s robustness.

Moreover, the choice of interpolation method plays a critical role in shaping attention distributions. \textbf{Bicubic interpolation}, while smooth, often over-smooths attention maps, making it less effective at capturing distinct object regions. \textbf{Nearest neighbor interpolation} demonstrates insensitivity to scaling factors above 1.0. \textbf{Area interpolation}, on the other hand, is overly sensitive to scaling factors below 1.0, resulting in attention maps that collapse onto irrelevant regions and fail to preserve essential object features. In contrast, \textbf{bilinear interpolation} achieves a balance between smoothness and precision, producing the most consistent and well-distributed attention maps. 

Thus, we adopt \textbf{bilinear interpolation} in our method to ensure optimal attention distribution and robust adversarial performance.

% \section{Examples on the 对抗图像}
% 在本节，我们可视化了我们的方法CWT在8个代理模型上生成的对抗图像在Resnet-18上的热图，如图\ref{fig:CWT}所示,我们能发现，CWT能够成功的迁移目标注意力到不同的地方

% \begin{figure*}[h]
% \begin{center}
% % \vspace{-1.2em}
%   \includegraphics[width=\textwidth]{figtex/figures/ggboy.png} %插入图片，[]中设置图片大小，{}中是图片文件
% \end{center}
% % \vspace{-1.2em}
% \caption{Heatmaps of the 对抗图像 由CWT在8个代理模型上生成}
% \label{fig:CWT}
% % \vspace{-1.5em}
% \end{figure*}

\section{Examples of Adversarial Images}
\begin{figure*}[h]
\begin{center}
  \includegraphics[width=0.9\textwidth]{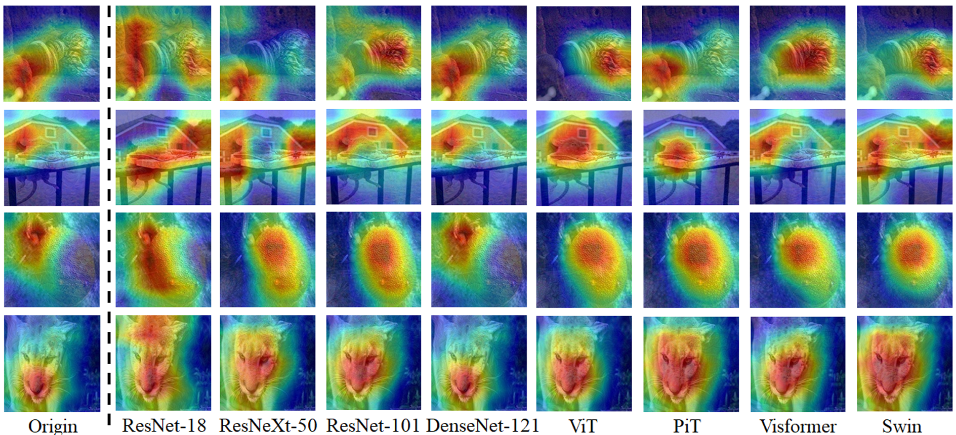} % Insert image
\end{center}
\caption{\textbf{Heatmaps of the adversarial images} generated by CWT on eight surrogate models. The heatmaps generated on ResNet-18.}
\label{fig:CWT}
\end{figure*}

In this section, we visualize the adversarial images generated by our method, CWT, on eight surrogate models, using heatmaps on ResNet-18, as shown in Figure \ref{fig:CWT}. It can be observed that CWT effectively transfers the target attention to different regions of the image.

% \newpage
% \input{sec/X_suppl}
% {
%     \small
%     \bibliographystyle{ieeenat_fullname}
%     \bibliography{appendix}
% }

\end{document}